\documentclass[10pt,twocolumn,letterpaper]{article}

\usepackage{cvpr}
\usepackage{times}
\usepackage{epsfig}
\usepackage{graphicx}
\usepackage{amsmath}
\usepackage{amssymb}
\usepackage[table]{xcolor}
\usepackage{subfig}
\usepackage{colortbl}
\usepackage{float}
\usepackage{afterpage}
\usepackage{multicol}
\usepackage[font=small]{caption}
\usepackage{subfloat}

\newcommand{\settitle}{\@maketitle}

\definecolor{Blue1}{RGB}{214, 235, 245}
\definecolor{Blue2}{RGB}{235, 245, 250}

\usepackage[pagebackref=true,breaklinks=true,letterpaper=true,colorlinks,bookmarks=false]{hyperref}

\cvprfinalcopy 

\def\cvprPaperID{7021} 

\ifcvprfinal\fi

\newcommand{\figref}[1]{figure~\ref{fig:#1}}
\newcommand{\secref}[1]{section~\ref{sec:#1}}

\begin{document}


\title{GQA: A New Dataset for Real-World Visual Reasoning \\ and Compositional Question Answering \\
\large \url{visualreasoning.net}} 


\author{Drew A. Hudson\\
Stanford University\\
353 Serra Mall, Stanford, CA 94305\\
{\tt\small dorarad@cs.stanford.edu}
\and
Christopher D. Manning\\
Stanford University\\
353 Serra Mall, Stanford, CA 94305\\
{\tt\small manning@cs.stanford.edu}
}

\maketitle

\begin{abstract}

We introduce GQA, a new dataset for real-world visual reasoning and compositional question answering, seeking to address key shortcomings of previous VQA datasets. We have developed a strong and robust question engine that leverages Visual Genome scene graph structures to create 22M diverse reasoning questions, which all come with functional programs that represent their semantics. We use the programs to gain tight control over the answer distribution and present a new tunable smoothing technique to mitigate question biases. Accompanying the dataset is a suite of new metrics that evaluate essential qualities such as consistency, grounding and plausibility. A careful analysis is performed for baselines as well as state-of-the-art models, providing fine-grained results for different question types and topologies. Whereas a blind LSTM obtains a mere 42.1\%, and strong VQA models achieve 54.1\%, human performance tops at 89.3\%, offering ample opportunity for new research to explore. We hope GQA will provide an enabling resource for the next generation of models with enhanced robustness, improved consistency, and deeper semantic understanding of vision and language.

\end{abstract}

\section{Introduction}
\label{sec:intro}

\thispagestyle{empty}

\begin{figure}
\centering
\subfloat{\includegraphics[width=\linewidth]{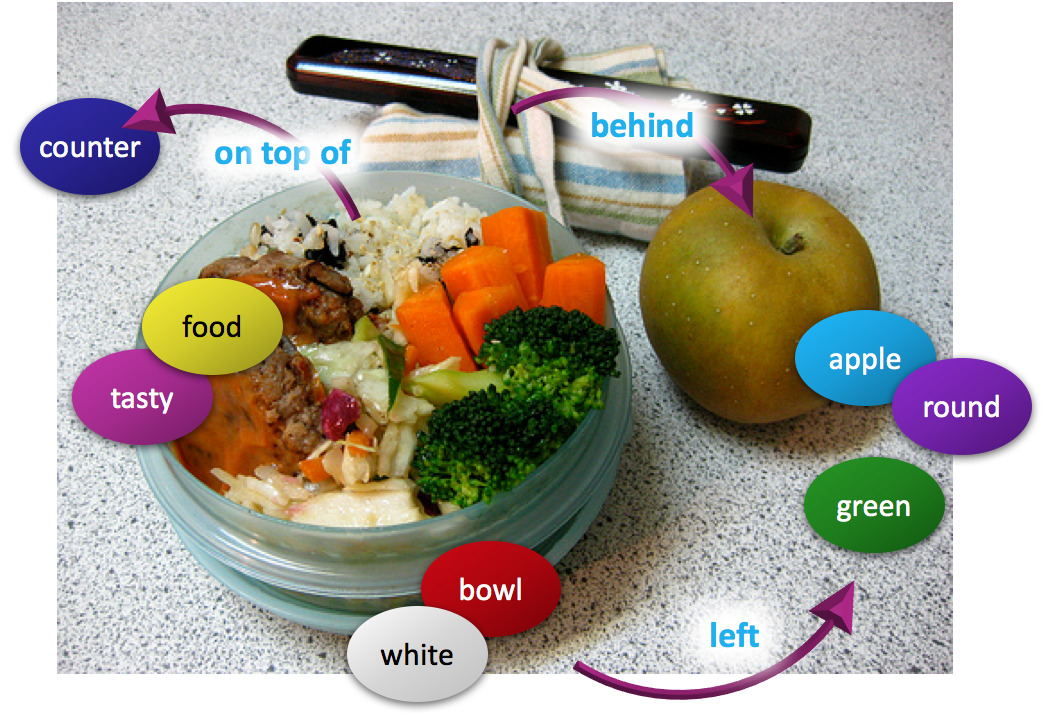}
}
\caption{\small Examples from the new GQA dataset for visual reasoning and compositional question answering:\\
\textit{Is the \textbf{\textcolor{magenta}{bowl}} to the right of the \textbf{\textcolor{teal}{green}} \textbf{\textcolor{blue}{apple}}?}\\
\textit{What type of \textbf{\textcolor{blue}{fruit}} in the image is \textbf{\textcolor{purple}{round}}?}\\
\textit{What color is the \textbf{\textcolor{blue}{fruit}} on the right side, red or \textbf{\textcolor{teal}{green}}?}\\
\textit{Is there any \textbf{\textcolor{violet}{milk}} in the \textbf{\textcolor{magenta}{bowl}} to the left of the \textbf{\textcolor{blue}{apple}}?}}
\label{plotss}
\end{figure}


It takes more than a smart guess to answer a good question. The ability to assimilate knowledge and use it to draw inferences is among the holy grails of artificial intelligence. A tangible form of this goal is embodied in the task of Visual Question Answering (VQA), where a system has to answer free-form questions by reasoning about presented images. 
The task demands a rich set of abilities as varied as object recognition, commonsense understanding and relation extraction, spanning both the visual and linguistic domains. In recent years, it has sparked substantial interest throughout the research community, becoming extremely popular across the board, with a host of datasets being constructed \cite{vqa1, vqa2, clevr, visual7w, vg} and numerous models being proposed \cite{bottomup, san, nmn, mcb, mac}. 

The multi-modal nature of the task and the diversity of skills required to address different questions make VQA particularly challenging. Yet, designing a good test that will reflect its full qualities and complications may not be that trivial. Despite the great strides that the field recently made, it has been established through a series of studies that existing benchmarks suffer from critical vulnerabilities that render them highly unreliable in measuring the actual degree of visual understanding capacities \cite{yinyang, vqa2, vqaanalysis, humanatt, cpvqa, revisiting, surveyMotivationStarve}. 



Most notable among the flaws of current benchmarks is the strong and prevalent real-world priors displayed throughout the data \cite{yinyang, vqa2, cpvqa} -- most tomatoes are red and most tables are wooden. These in turn are exploited by VQA models, which become heavily reliant upon such statistical biases and tendencies within the answer distribution to largely circumvent the need for true visual scene understanding \cite{vqaanalysis, vqa2, clevr, humanatt}. This situation is exacerbated by the simplicity of many of the questions, from both linguistic and semantic perspectives, which in practice rarely require much beyond object recognition \cite{nlvr2}. Consequently, early benchmarks led to an inflated sense of the state of  scene understanding, severely diminishing their credibility \cite{biases}. Aside from that, the lack of annotations regarding question structure and content leaves it difficult to understand the factors affecting models' behavior and performance and to identify the root causes behind their mistakes.
To address these shortcomings, while retaining the visual and semantic richness of real-world images, we introduce GQA, a new dataset for visual reasoning and compositional question answering. We have developed and carefully refined a robust \textit{question engine}, leveraging \textbf{content}: information about objects, attributes and relations provided through Visual Genome \textit{Scene Graphs} \cite{vg}, along with \textbf{structure}: a newly-created extensive linguistic grammar which couples hundreds of structural patterns and detailed lexical semantic resources. Together, they are combined in our engine to generate over 22 million novel and diverse questions, which all come with structured representations in the form of functional programs that specify their contents and semantics, and are visually grounded in the image scene graphs.

GQA questions involve varied reasoning skills, and multi-step inference in particular. We further use the associated semantic representations to greatly reduce biases within the dataset and control for its question type composition, downsampling it to create a 1.7M balanced dataset. Contrary to VQA 2.0, here we balance not only binary questions, but also open ones, by applying a tunable smoothing technique that makes the answer distribution for each question group more uniform. Just like a well-designed exam, our benchmark makes the educated guesses strategy far less rewarding, and demands instead more refined comprehension of both the visual and linguistic contents. 





Along with the dataset, we have designed a suite of new metrics, which include consistency, validity, plausibility, grounding and distribution scores, to complement the standard accuracy measure commonly used in assessing methods' performance. Indeed, studies have shown that the accuracy metric alone does not account for a range of anomalous behaviors that models demonstrate, such as ignoring key question words or attending to irrelevant image regions \cite{vqaanalysis, humanatt}. Other works have argued for the need to devise new evaluation measures and techniques to shed more light on systems' inner workings \cite{surveyMotivationStarve, tips, graph, tmpRochester}. In fact, beyond providing new metrics, GQA can even directly support the development of more interpretable models, as it provides a sentence-long explanation that corroborates each answer, and further associates each word from both the questions and the responses with a visual pointer to the relevant region in the image, similar in nature to datasets by Zhu \etal \cite{visual7w}, Park \etal \cite{exp1}, and Li \etal \cite{exp2}. These in turn can serve as a strong supervision signal to train models with enhanced transparency and accessibility.

GQA combines the best of both worlds, having clearly defined and crisp semantic representations on the one hand but enjoying the semantic and visual richness of real-world images on the other. 
Our three main contributions are (1)~the GQA dataset as a resource for studying visual reasoning; (2)~development of an effective method for generating a large number of semantically varied questions, which marries scene graph representations with computational linguistic methods; (3)~new metrics for GQA, that allow for better assessment of system success and failure modes, as demonstrated through a comprehensive performance analysis of existing models on this task. We hope that the GQA dataset will provide fertile ground for the development of novel methods that push the boundaries of question answering and visual reasoning.



\section{Related Work}
\label{sec:related}

\begin{figure*}[t]
\begin{center}
\subfloat{\includegraphics[width=0.97\linewidth]{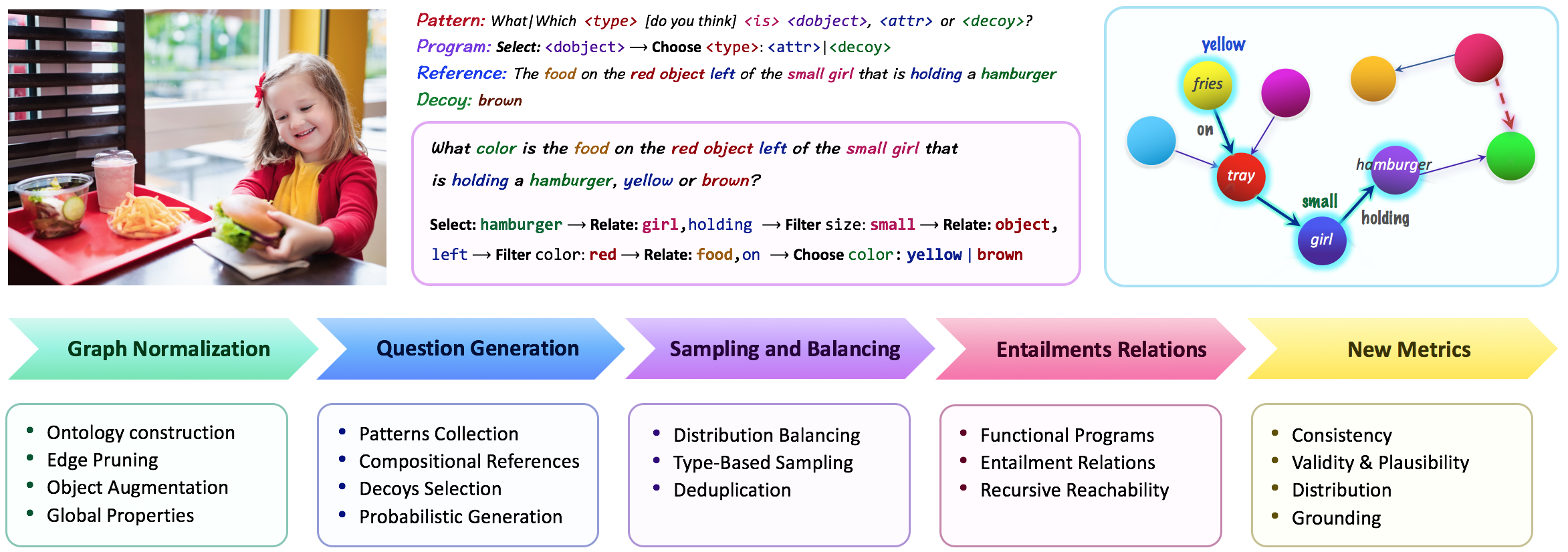}}
\end{center}
   \caption{Overview of the GQA construction process. Given an image annotated with a scene graph of its objects, attributes and relations, we produce compositional questions by traversing the graph. Each question has both a standard natural-language form and a functional program representing its semantics. Please refer to \secref{dataset}  for further detail.}
\label{fig:process}
\end{figure*}
Recent years have witnessed tremendous progress in visual understanding. Multiple attempts have been made to mitigate the systematic biases of VQA datasets as discussed in \secref{intro} \cite{vqa2, yinyang, cpvqa, clevr}, but they fall short in providing an adequate solution: Some approaches operate over constrained and synthetic images \cite{yinyang, clevr}, neglecting the realism and diversity natural photos provide. Meanwhile, Goyal \etal \cite{vqa2} associate most of the questions in VQA1.0 with a pair of similar pictures that result in different answers. While offering partial relief, this technique fails to address open questions, leaving their answer distribution largely unbalanced. In fact, since the method does not cover 29\% of the questions due to limitations of the annotation process, even within the binary ones biases still remain.\footnote{For VQA1.0, blind models achieve 50\% in accuracy without even considering the images whatsoever \cite{vqa1}. Similarly, for VQA2.0, 67\% and 27\% of the binary and open questions respectively are answered correctly by such models \cite{vqa2}.}

At the other extreme, Agrawal \etal \cite{cpvqa} partition the questions into training and validation sets such that their respective answer distributions become intentionally dissimilar. While undoubtedly challenging, these adversarial settings penalize models, maybe unjustly, for learning salient properties of the training data. In the absence of other information, making an educated guess is a legitimate choice -- a valid and beneficial strategy pursued by machines and people alike \cite{humanguess1, humanguess2, humanguess3}. What we essentially need is a balanced test that is more resilient to such gaming strategies, as we strive to achieve with GQA.



In creating GQA, we drew inspiration from the CLEVR task \cite{clevr}, which consists of compositional questions over synthetic images. However, its artificial nature and  low diversity, with only a handful of object classes and properties, makes it particularly vulnerable to memorization of all combinations, thereby reducing its effective degree of compositionality. Conversely, GQA operates over real images and a large semantic space, making it much more challenging. Even though our questions are not natural as in other VQA datasets \cite{vqa2, visual7w}, they display a rich vocabulary and diverse linguistic and grammatical structures. They may serve in fact as a cleaner benchmark to asses models in a more controlled and comprehensive fashion, as discussed below. 
The task of question generation has been explored in earlier work, mostly for the purpose of data augmentation. Contrary to GQA, those datasets are either small in scale \cite{tmpDaquar} or use only a restricted set of objects and a handful of non-compositional templates \cite{tmpRochester, tmpPremise}. Neural alternatives to visual question generation have been recently proposed \cite{vqg1, vqg2, vqg5}, but they aim at a quite different goal of creating engaging but potentially inaccurate questions about the wider context of the image such as subjective evoked feelings or speculative events that may lead to or result from  the depicted scenes \cite{vqg1}. 



\section{The GQA Dataset}
\label{sec:dataset}

\begin{figure}
\vspace*{-2mm}
\centering
\subfloat{\includegraphics[width=0.48\linewidth]{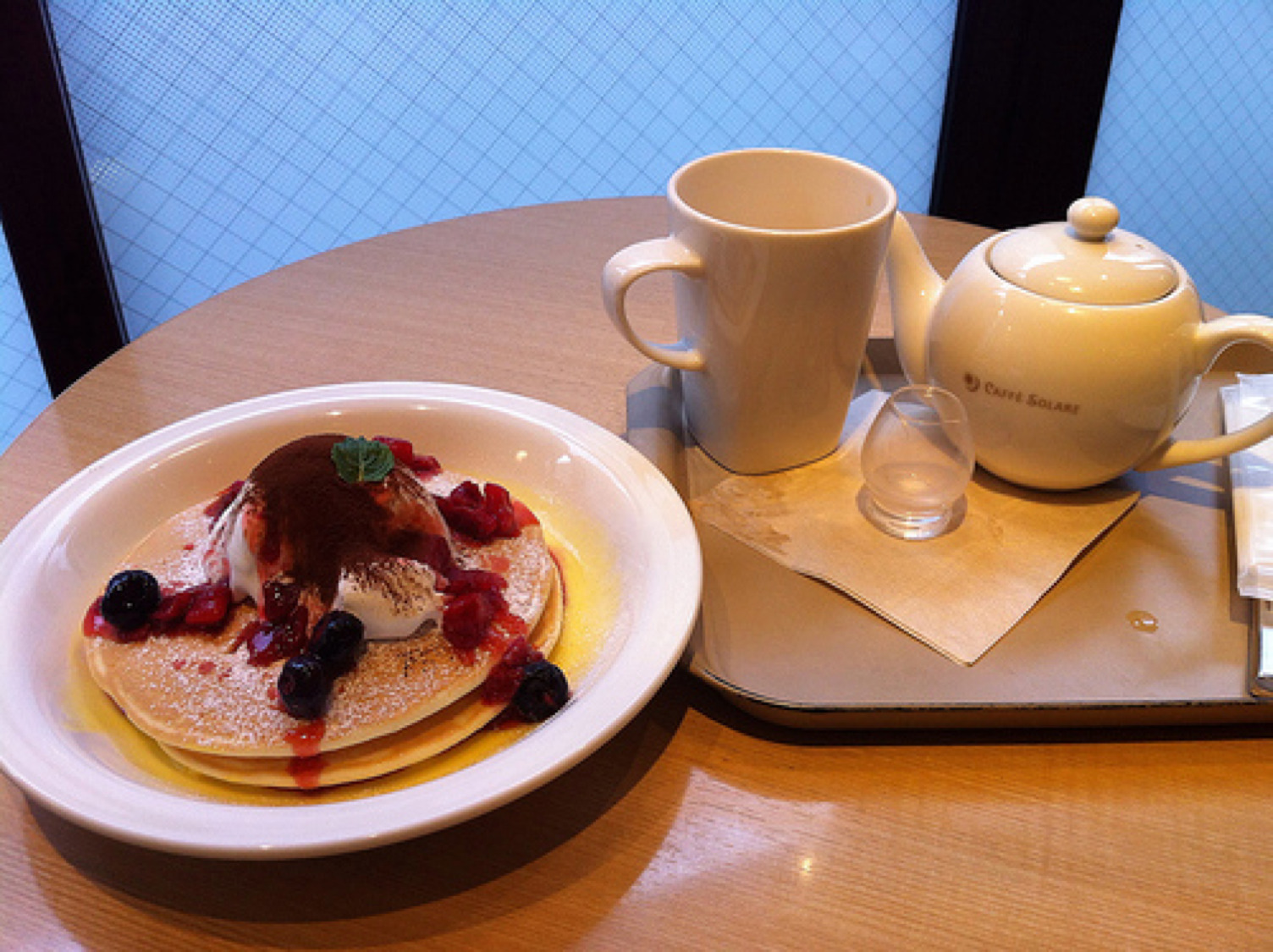}}
\hfill
\subfloat{\includegraphics[width=0.48\linewidth]{imgs/img2.jpg}}

\footnotesize\raggedright \textbf{A1}. Is the \textbf{{\color{teal}tray}} on top of the \textbf{{\color{violet}table}} black or light brown? light brown \\
\textbf{A2}. Are the \textbf{{\color{magenta}napkin}} and the \textbf{{\color{orange}cup}} the same color? yes \\
\textbf{A3}. Is the small \textbf{{\color{violet}table}} both oval and wooden? yes \\
\textbf{A4}. Is there any \textbf{{\color{olive}fruit}} to the left of the \textbf{{\color{teal}tray}} the \textbf{{\color{orange}cup}} is on top of? yes \\
\textbf{A5}. Are there any \textbf{{\color{orange}cups}} to the left of the \textbf{{\color{teal}tray}} on top of the \textbf{{\color{violet}table}}? no \\

\textbf{B1}. What is the brown \textbf{{\color{cyan}animal}} sitting inside of? \textbf{{\color{purple}box}} \\
\textbf{B2}. What is the large \textbf{{\color{magenta}container}} made of? cardboard \\
\textbf{B3}. What \textbf{{\color{cyan}animal}} is in the \textbf{{\color{purple}box}}? \textbf{{\color{cyan}bear}} \\
\textbf{B4}. Is there a \textbf{{\color{teal}bag}} to the right of the green \textbf{{\color{orange}door}}? no \\
\textbf{B5}. Is there a \textbf{{\color{magenta}box}} inside the plastic \textbf{{\color{teal}bag}}? no 

\caption{Examples of questions from the GQA dataset.}
\label{fig:examples}
\end{figure}

The GQA dataset centers around real-world reasoning, scene understanding and compositional question answering. It consists of 113K images and 22M questions of assorted types and varying compositionality degrees, measuring performance on an array of reasoning skills such as object and attribute recognition, transitive relation tracking, spatial reasoning, logical inference and comparisons. Figure \ref{fig:process} provides a brief overview of the GQA components and generation process, and figure \ref{fig:examples} presents multiple instances from the dataset. The dataset along with further information are available at \url{visualreasoning.net}.





The images, questions and corresponding answers are all accompanied by matching semantic representations: Each \textit{image} is annotated with a dense \textbf{Scene Graph} \cite{sg, vg}, representing the objects, attributes and relations it contains. Each \textit{question} is associated with a \textbf{functional program} which lists the series of reasoning steps needed to be performed to arrive at the answer. Each \textit{answer} is augmented with both textual and visual justifications, pointing to the relevant region within the image. 

The structured representations and detailed annotations for images and questions offer multiple advantages. They enable tight control over the answer distribution, which allows us to create a balanced set of challenging questions, and support the formulation of a suite of new metrics that aim to provide deeper insight into models' behavior. They  facilitate performance assessment along various axes of question type and topology, and may open the door for the development of novel methods with more grounded and transparent knowledge representation and reasoning.
We proceed by describing the GQA question engine and the four-step dataset construction pipeline: First, we thoroughly clean, normalize, consolidate and augment the Visual Genome scene graphs \cite{vg} linked to each image. Then, we traverse the objects and relations within the graphs, and marry them with grammatical patterns gleaned from VQA 2.0 \cite{vqa2} and sundry probabilistic grammar rules to produce a semantically-rich and diverse set of questions. In the third stage, we use the underlying semantic forms to reduce biases in the conditional answer distribution, resulting in a balanced dataset that is more robust against shortcuts and guesses. Finally, we discuss the question functional representation, and explain how we use it to compute entailment between questions, supporting new evaluation metrics. 



\subsection{Scene Graph Normalization}
Our starting point in creating the GQA dataset is the Visual Genome Scene Graph annotations \cite{vg} that cover 113k images from COCO \cite{coco} and Flickr \cite{flickr}.\footnote{We extend Visual Genome dataset with 5k hidden scene graphs collected through crowdsourcing, used for the test set.} The scene graph serves as a formalized representation of the image: each node denotes an \textbf{object}, a visual entity within the image, like a person, an apple, grass or clouds. It is linked to a bounding box specifying its position and size, and is marked up with about 1--3 \textbf{attributes}, properties of the object: e.g., its color, shape, material or activity. The objects are connected by \textbf{relation} edges, representing actions (verbs), spatial relations (prepositions), and comparatives. 

The scene graphs are annotated with free-form natural language. In order to use them for question generation we first have to normalize the graphs and their vocabulary. We provide here a brief overview of the normalization process, and present a more detailed description in the supplementary. First, we create a clean, consolidated and unambiguous ontology over the graph with 2690 classes including various objects, attributes and relations. We further augment it with semantic and linguistic information which will aid us in creating grammatical questions. Then, we prune inaccurate or unnatural edges, using combination of object detection confidences, n-gram frequencies, co-occurrence statistics, word embedding distances, category-based rules, and manual curation. Finally, we enrich the graph with positional information (absolute and relative) as well as semantic properties (location, weather). By the end of this stage, the resulting scene graphs have clean, unified, rich and unambiguous semantics for both the nodes and the edges.

\subsection{The Question Engine}
At the heart of our pipeline is the question engine, responsible for producing diverse, relevant and grammatical questions with varying degrees of compositionality. The generation process harnesses two resources: one is the scene graphs which fuel the engine with rich content -- information about objects, attributes and relationships; the other is the structural patterns, a mold that shapes the content, casting it into a question. 

Our engine operates over 524 patterns, spanning 117 question groups, and 1878 answers which are based on the scene graphs. Each group is associated with three components: (1) a functional program that represents its semantics; (2) A set of textual rephrases which express it in natural language, e.g., ``What{\textbar}Which \texttt{<type>} [do you think] \texttt{<is>} \texttt{<theObject>}?"; (3) A pair of short and long answers: e.g., \texttt{<attribute>} and ``The \texttt{<object>} \texttt{<is>} \texttt{<attribute>}." respectively.\footnote{Note that the long answers can serve as textual justifications, especially for questions that require increased reasoning such as logical inference, where a question like \textit{``Is there a red apple in the picture?"} may have the answer: \textit{``No, there is an apple, but it is green"}} 

We begin from a seed set of 250 manually constructed patterns, and extend it with 274 natural patterns derived from VQA1.0 \cite{vqa1} through templatization of words from our ontology.\footnote{For instance, a question-answer pair in VQA1.0 such as \textit{``What color is the apple? red"} turns after templatization into ``What \texttt{<type>} \texttt{<is>} the \texttt{<object>}? \texttt{<attribute>}".} To increase the question diversity, apart from using synonyms for objects and attributes, we incorporate probabilistic sections into the patterns, such as optional phrases \textit{[x]} and alternate expressions (\textit{x{\textbar}y}), which get instantiated at random. 

{\em It is important to note that the patterns do not strictly limit the structure or depth of each question, but only outline their high-level form}, as many of the template fields can be populated with nested compositional references. For instance, in the pattern above, we may replace \texttt{<theObject>} with ``\texttt{the apple to the left of the white refrigerator}". 

To achieve that compositionality, we compute for each object a set of candidate references, which can either be \textbf{direct}, \eg \textit{the bear}, \textit{this animal}, or \textbf{indirect}, using modifiers, \eg \textit{the white bear}, \textit{the bear on the left}, \textit{the animal behind the tree}, \textit{the bear that is wearing a coat}. Direct references are used when the uniqueness of the object can be confidently confirmed by object detectors, making the corresponding references unambiguous. Alternatively, we use indirect references, leading to multi-step questions as varied as \textit{Who is looking at the animal that is wearing the red coat in front of the window?}, and thus greatly increasing the patterns' effective flexibility. This is the key ingredient behind the automatic generation of compositional questions.
Finally, we compute a set of decoys for the scene graph elements. Indeed, some questions, such as negative ones or those that involve logical inference, pertain to the absence of an object or to an incorrect attribute. Examples include \textit{Is the apple green?} for a red apple, or \textit{Is the girl eating ice cream?} when she is in fact eating a cake. Given a triplet ${(s, r , o)}$, (\eg \textit{(girl, eating, cake}) we select a distractor ${\hat{o}}$ considering its likelihood to be in relation with \textit{s} and its plausibility to co-occur in the context of the other objects in the depicted scene. A similar technique is applied in selecting attribute decoys (\eg a green apple). While choosing distractors, we exclude from consideration candidates that we deem too similar (\eg \textit{pink} and \textit{orange}), based on a manually defined list for each concept in the ontology.

\begin{figure}
\centering
\subfloat{\includegraphics[width=\linewidth]{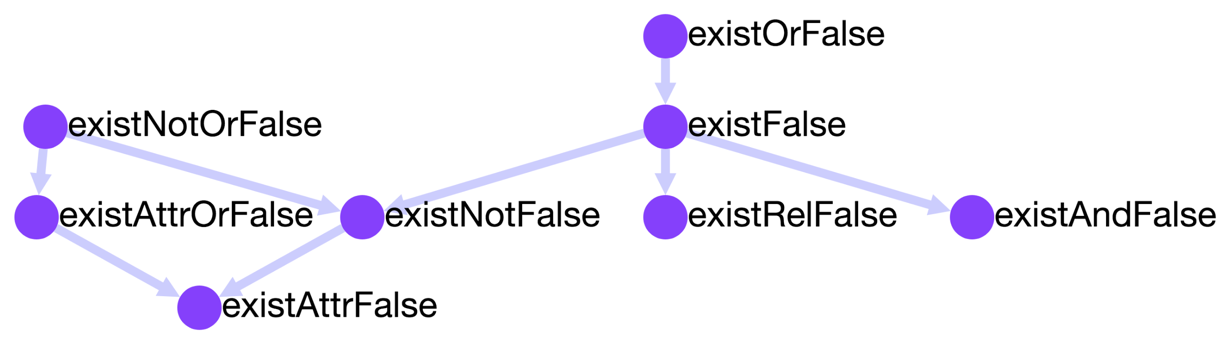}}
\caption{Examples of entailment relations between different question types. Refer to \secref{exppp} for further detail.}
\label{fig:entailSmall}
\end{figure}

Having all resources prepared: (1) the clean scene graphs, (2) the structural patterns, (3) the object references and (4) the decoys, we can proceed to generating the questions! We traverse the graph, and for each object, object-attribute pair or subject-relation-object triplet, we produce relevant questions by instantiating a randomly selected question pattern, \eg \textit{``What \texttt{<type>} is \texttt{<theObject>}, \texttt{<attribute>} or \texttt{<cAttribute>}?"}, populating all the fields with the matching information, yielding, for example, the question: \textit{``What (color) (is) the (apple on the table), (red) or (green)?"}. When choosing object references, we avoid selecting those that disclose the answer or repeat information, \eg \textit{``What color is the red apple?"} or \textit{``Which dessert sits besides the apple to the left of the cake?"}. We also avoid asking about relations that tend to have multiple instances for the same object, \eg asking what object is on the table, as there may be multiple valid answers.

By the end of this stage, we obtain a diverse set of 22M interesting, challenging and grammatical questions, pertaining to each and every aspect of the image.










\subsection{Functional Representation and Entailment}
\label{sec:exppp}

\begin{figure*}
\subfloat{\includegraphics[width=0.355\linewidth]{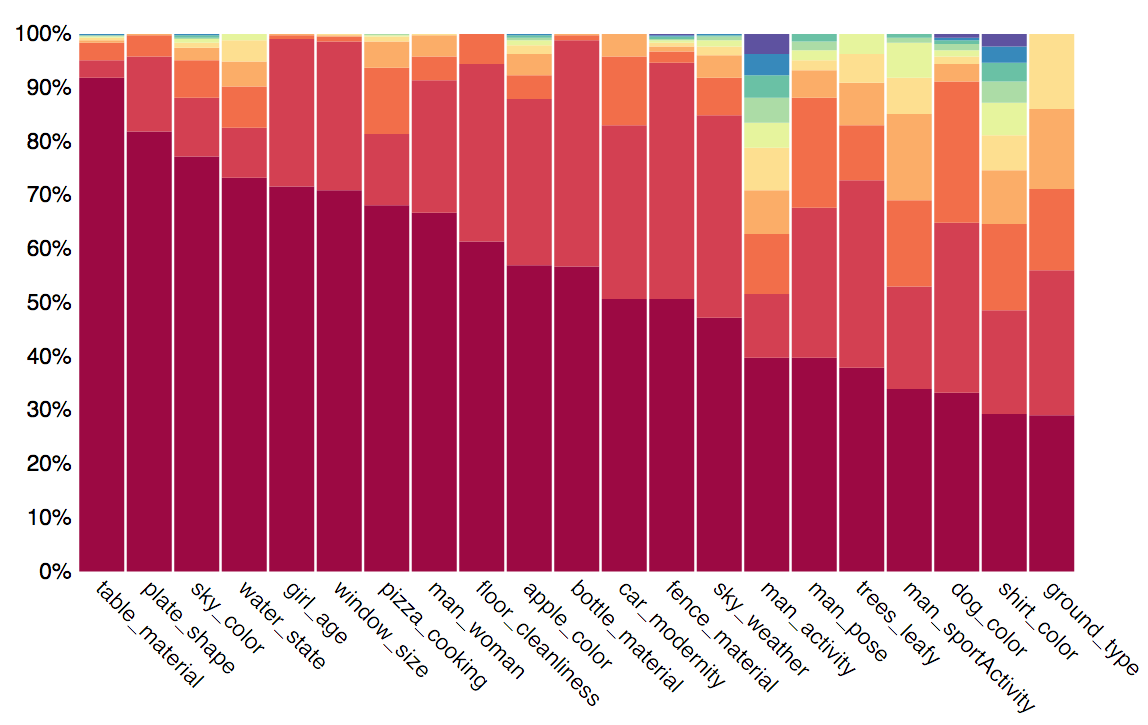}}
\hspace{1mm}
\subfloat{\includegraphics[width=0.355\linewidth]{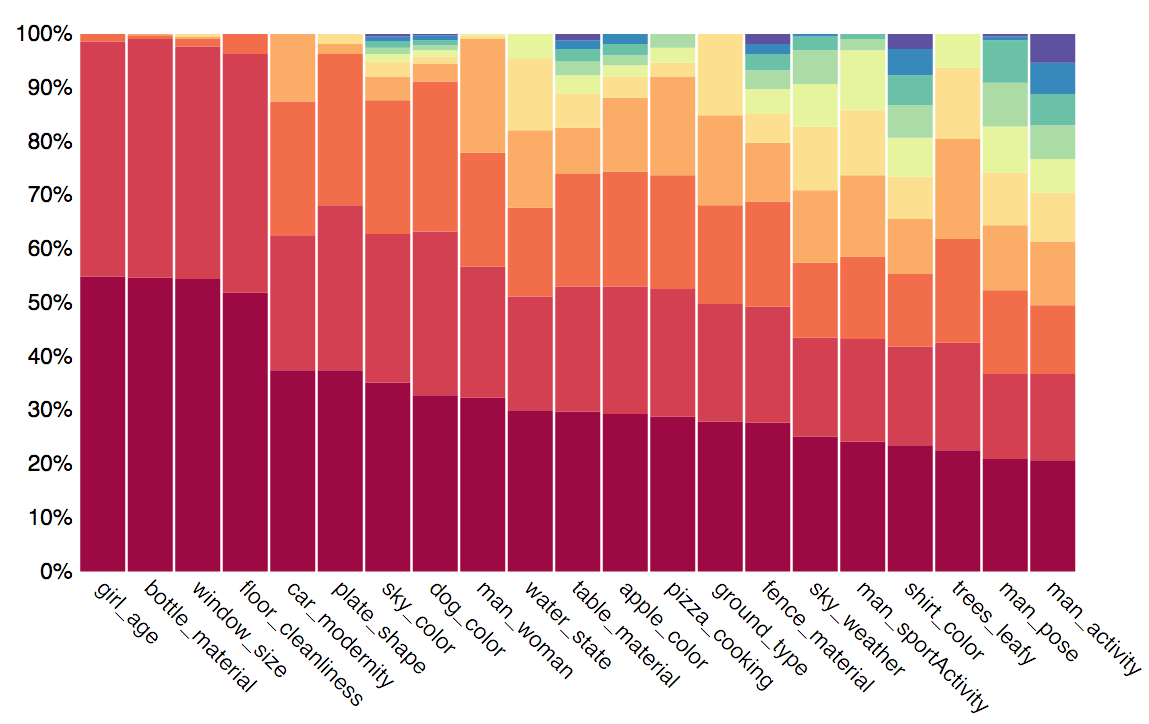}}
\hspace{2mm}
\subfloat{\includegraphics[width=0.235\linewidth]{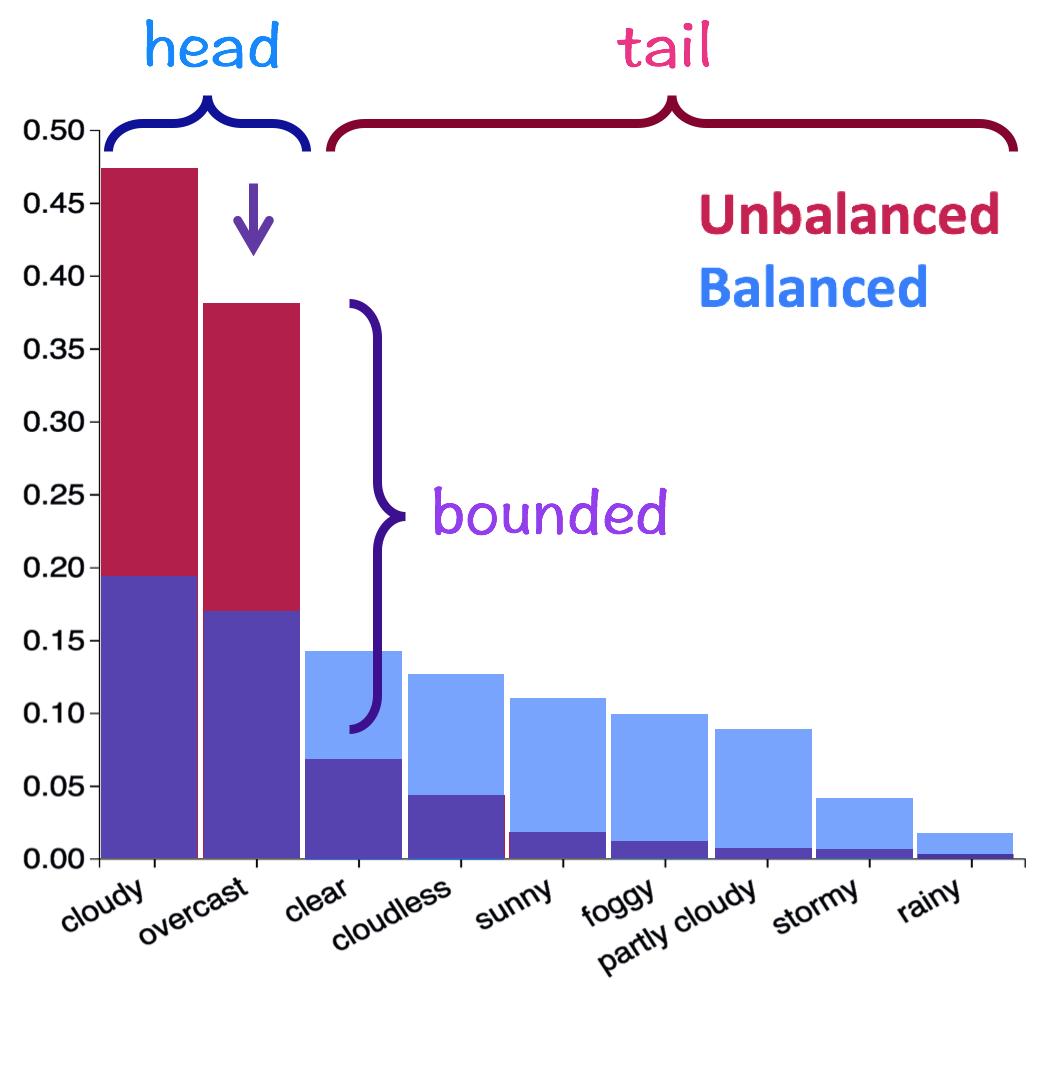}}

   \caption{Visualization of the balancing process. The conditional answer distribution before (\textbf{left}) and after (\textbf{middle}) the balancing for a selection of question groups. We show the top 10 answers, where the column height corresponds to the relative frequency of each answer. We can see that on the left the distributions are heavily biased, while on the middle it is more uniform and with heavier tails, while intentionally retaining the original real-world tendencies up to a tunable degree. \textbf{Right}: An illustration of the balancing process.}
   
   
\label{fig:balancing}
\end{figure*}

Each question pattern is associated with a structured representation in the form of a functional program. For instance, the question \textit{What color is the apple on the white table?} is semantically equivalent to the following program: \textit{select:} {table} $\rightarrow$ \textit{filter:} {white} $\rightarrow$ \textit{relate(subject,on):} {apple} $\rightarrow$ \textit{query:} {color}. As we can see, these programs are composed of atomic operations such as object selection, traversal along a relation edge, or an attribute verification, which are then chained together to create challenging reasoning questions. 

The semantically unambiguous representations offer multiple advantages over free-form unrestricted questions. For one thing, they enable comprehensive assessment of methods by dissecting their performance along different axes of question textual and semantic lengths, type and topology, thus facilitating the diagnosis of their success and failure modes (\secref{exps}). Second, they aid us in balancing the dataset distribution, mitigating its question-conditional priors and guarding against educated guesses (\secref{balancing}). Finally, they allow us to identify entailment and equivalence relations between different questions: knowing the answer to the question \textit{What color is the apple?} allows a coherent learner to infer the answer to the questions \textit{Is the apple red? Is it green?} etc. The same goes especially for questions that involve logical inference like \textit{or} and \textit{and} operations or spatial reasoning, \eg \textit{left} and \textit{right}. 

As further discussed in \secref{metrics}, this entailment property can be used to measure the coherence and consistency of the models, shedding new light on their inner workings, compared to the widespread but potentially misleading accuracy metric. We define direct entailment relations between the various functional programs and use these to recursively compute all the questions that can be entailed from a given source. A complete catalog of the functions, their associated question types, and the entailment relations between them is provided in the supplementary. 
 
\subsection{Sampling and Balancing}
\label{sec:balancing}
One of the main issues of existing VQA datasets is the prevalent question-conditional biases that allow learners to make educated guesses without truly understanding the presented images, as explained in \secref{intro}. However, precise representation of the question semantics can allow tighter control over these biases, having the potential to greatly alleviate the problem. We leverage this observation and use the functional programs attached to each question to smooth out the answer distribution.

Given a question's functional program, we derive two labels, global and local: The global label assigns the question to its answer type, \eg \textit{color} for \textit{What color is the apple?}. The local label further considers the main subject/s of the question, \eg. \textit{apple-color} or \textit{table-material}. We use these labels to partition the questions into groups, and smooth the answer distribution of each group within the two levels of granularity, first globally, and then locally. 

For each group, we first compute its answer distribution $P$ for each group, which we then downsample (formally, using rejection-sampling) to fit a smoother answer distribution $Q$ derived through the following procedure: We iterate over the answers of that group in decreasing frequency order, and reweight $P$'s head up to the current iteration to make it more comparable to the tail size. While repeating this operation as we go through the answers, iteratively ``moving" probability from the head into the tail \cite{emd}, we also maintain minimum and maximum ratios between each pair of subsequent answers (sorted by frequency). This ensures that the relative frequency-based answer ranking stays the same. 

The advantage of this scheme is that it retains the  general real-world tendencies, smoothing them out up to a tunable degree to make the benchmark more challenging and less biased. Refer to \figref{balancing} for a visualization and to the supplementary for a precise depiction of the procedure. Since balancing is performed in two granularity levels, the obtained answer distributions are made more uniform both locally and globally. Quantitatively, the entropy of the answer distribution is increased by 72\%, confirming the success of this stage.

Finally, we downsample the questions based on their type to control the dataset type composition, and filter out redundant questions that are too semantically similar to existing ones. We split the dataset into 70\% train, 10\% validation, 10\% test and 10\% challenge, making sure that all the questions about a given image appear in the same split.

\section{Analysis and Baseline Experiments}

In the following, we provide an analysis of the GQA dataset and evaluate the performance of baselines, state-of-the-art models and human subjects, revealing a large gap from the latter. To establish the diversity and realism of GQA questions, we test transfer performance between the GQA and VQA datasets. We then introduce the new metrics that complement our dataset, present quantitative results and discuss their implications and merits. In the supplementary, we perform a head-to-head comparison between GQA and the popular VQA 2.0 dataset \cite{vqa2}, and proceed with further diagnosis of the current top-performing model, MAC \cite{mac}, evaluating it along multiple axes such as training-set size, question length and compositionality degree.   


\subsection{Dataset Analysis and Comparison}
\label{sec:analysis}


 
\begin{figure}
\vspace*{-4.2mm}
\centering
\subfloat{\includegraphics[width=0.33\linewidth]{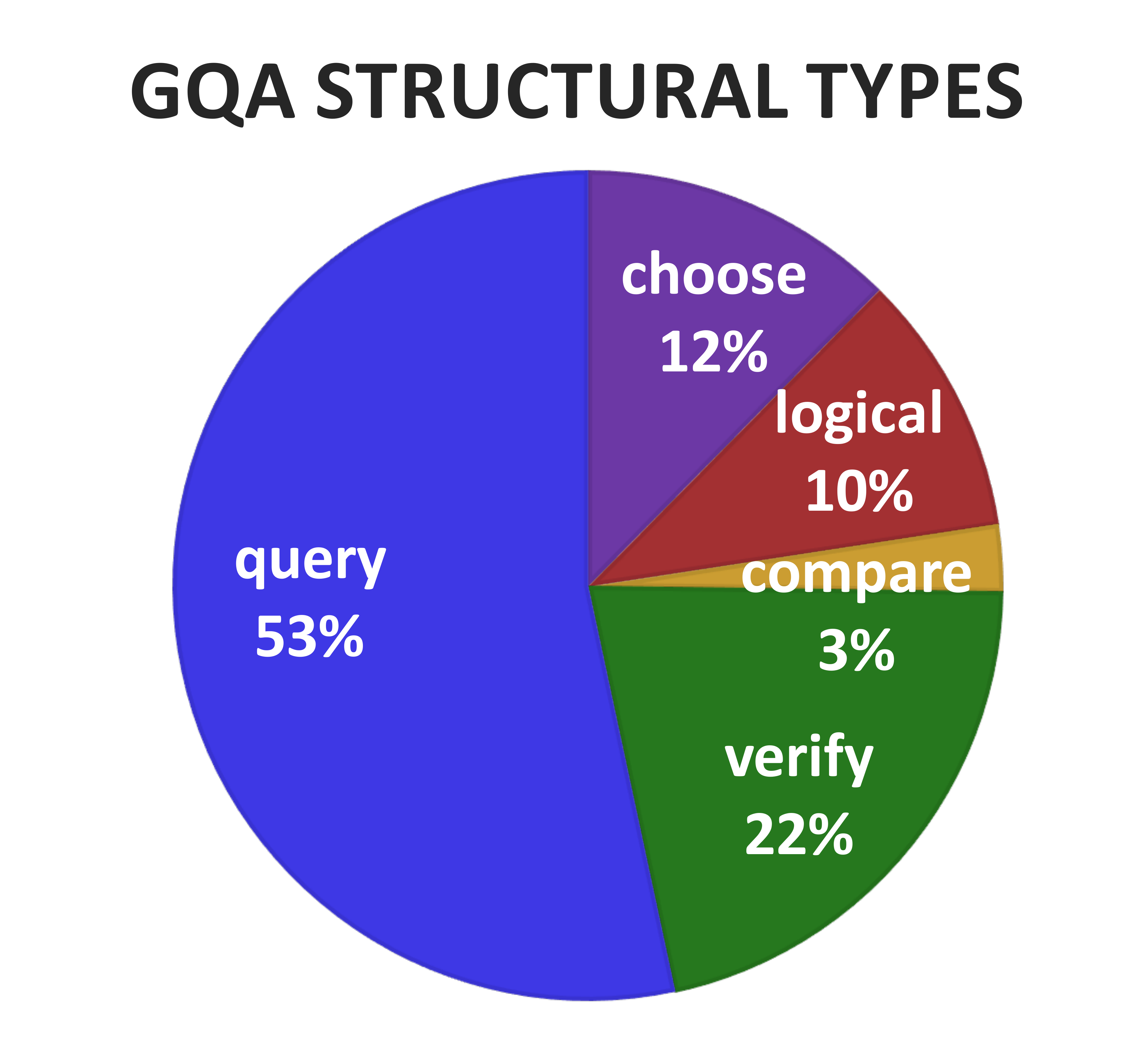}}
\subfloat{\includegraphics[width=0.33\linewidth]{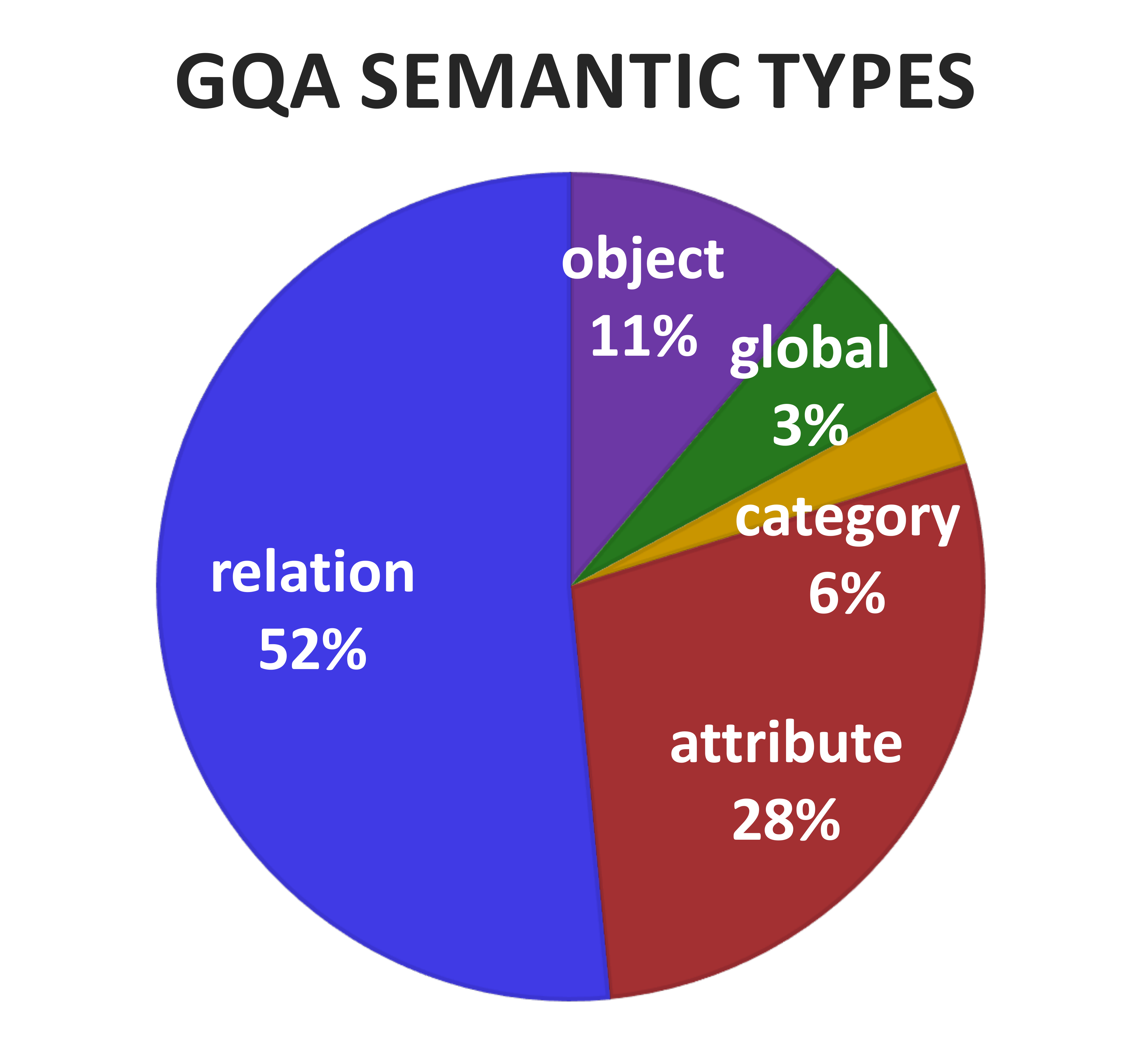}}
{\vspace*{-0.2mm}
\subfloat{\includegraphics[width=0.33\linewidth]{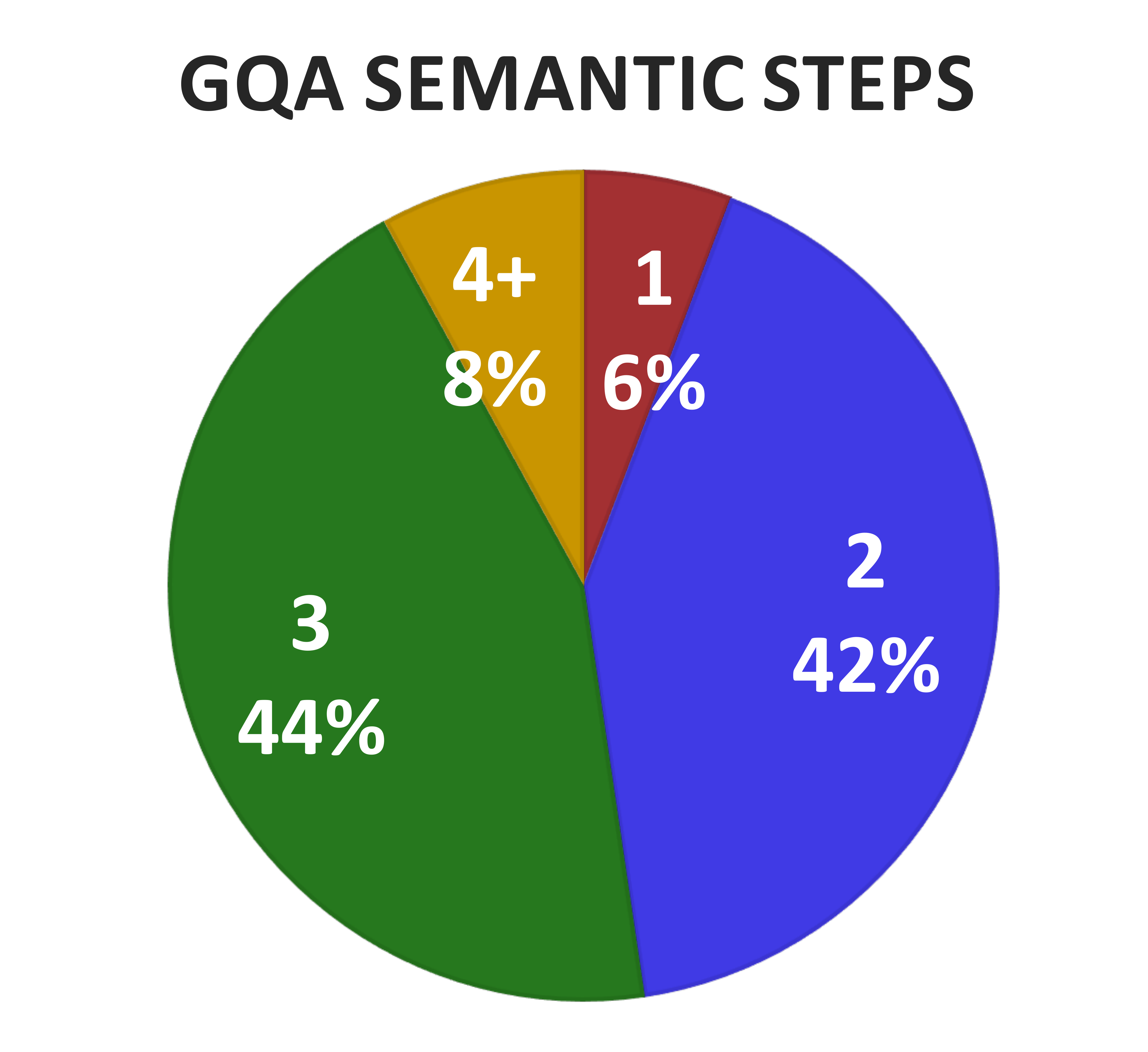}}}

\subfloat{\includegraphics[width=\linewidth]{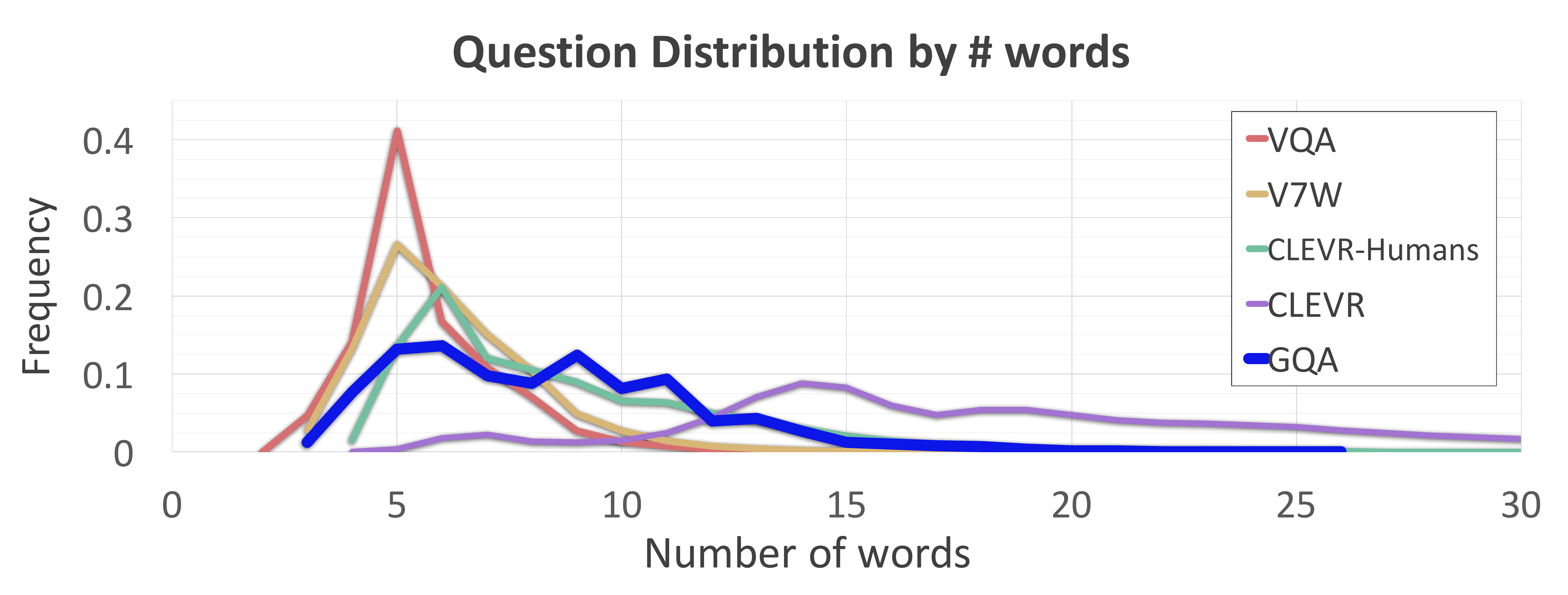}}

\caption{\textbf{Top}: Dataset statistics, partitioned into structural types, semantic types, and the number of reasoning steps. \textbf{Bottom}: VQA datasets question length distribution.}
\label{fig:cakes}
\label{fig:distt}
\end{figure}

The GQA dataset consists of 22,669,678 questions over 113,018 images, which cover wide range of reasoning skills and vary in length and number of required inference-steps (\figref{cakes}). The dataset has a vocabulary size of 3097 words and 1878 possible answers. While smaller than natural language datasets, further investigation reveals that it covers 88.8\% and 70.6\% of VQA questions and answers respectively, corroborating its wide diversity. A wide selection of dataset visualizations is provided in the supplementary.

We associate each question with two types: structural and semantic. The \textbf{structural type} is derived from the final operation in the question's functional program. It can be (1) \textbf{verify} for yes/no questions, (2) \textbf{query} for all open questions, (3) \textbf{choose} for questions that present two alternatives to choose from, \eg \textit{``Is it red or blue?";} (4) \textbf{logical} which involve logical inference, and (5) \textbf{compare} for comparison questions between two or more objects. The \textbf{semantic type} refers to the main subject of the question: (1) \textit{object}: for existence questions, (2) \textit{attribute}: consider the properties or  position of an object, (3) \textit{category}: related to object identification within some class, (4) \textit{relation}: for questions asking about the subject or object of a described relation (\eg \textit{``what is the girl wearing?"}), and (5) \textit{global}: about overall properties of the scene such as weather or place. As shown in \figref{cakes}, the questions' types vary at both the semantic and structural levels.

\subsection{Baseline Experiments}
\label{sec:exps}



We analyze an assortment of models as well as human subjects on GQA. The evaluation results are shown in table \ref{table2}. Baselines include a ``blind" LSTM model with access to the questions only, a ``deaf" CNN model with access to the images only, an LSTM+CNN model, and two prior models based on the question group, local or global, which return the most common answer for each group, as defined in \secref{exppp}. We can see that they all achieve low results of 17.82\%--41.07\%. For the LSTM model, inspection of specific question types reveals that it achieves only 22.7\% for open \textit{query} questions, and not far above chance for binary question types. We also evaluate the performance of the bottom-up attention model \cite{bottomup} -- the winner of the 2017 VQA challenge, and the MAC model \cite{mac} -- a state-of-the-art compositional attention model for CLEVR \cite{clevr}. While surpassing the baselines, they still perform well below human scores\footnote{To evaluate human performance, we used Amazon Mechanical Turk to collect human responses for 4000 random questions, taking the majority over 5 answers per question.}, offering ample opportunity for further research in the visual reasoning domain.


 





\subsection{Transfer Performance}
We tested the transfer performance between the GQA and VQA datasets, training on one and testing on the other: A MAC model trained on GQA achieves 52.1\% on VQA before fine-tuning and 60.5\% afterwards. Compare these with 51.6\% for LSTM+CNN and 68.3\% for MAC, when both are trained and tested on VQA. These quite good results demonstrate the realism and diversity of GQA questions, showing that the dataset can serve as a good proxy for human-like questions. In contrast, MAC trained on VQA gets 39.8\% on GQA before fine-tuning and 46.5\% afterwards, illustrating the further challenge GQA poses.

\subsection{New Evaluation Metrics}
\label{sec:metrics}

\begin{table*}
\centering
\scriptsize 
\begin{tabular}{lcccccccc}
\rowcolor{Blue1}
\scriptsize \textbf{Metric} & \textbf{Global Prior} & \textbf{Local Prior} & \textbf{CNN} & \textbf{LSTM} & \textbf{CNN+LSTM} & \textbf{BottomUp} & \textbf{MAC} & \textbf{Humans} \\ 
Open & 
16.52 &
16.99 &
1.74	 &
22.69 &
31.80 &
34.83 &
38.91 & 87.4 \\[0.2em] 

Binary &
42.99 &
47.53 &
36.05 &
61.90 &
63.26 &
66.64 &
71.23 & 91.2 \\[0.2em] 

\rowcolor{Blue2}
Query &
16.52 &
16.99 &
1.55 &
22.69 &
31.80 &
34.83 &
38.91 & 87.4 \\[0.2em] 

\rowcolor{Blue2}
Compare &
35.59 &
41.91 &
36.34 &
57.79 &
56.62 &
56.32 &
60.04 & 93.1\\[0.2em] 

\rowcolor{Blue2}
Choose &
17.45 &
26.58 &
0.85 &
57.15 &
61.40 &
66.56 &
70.59 & 94.3\\[0.2em] 

\rowcolor{Blue2}
Logical &
50.32 &
50.11 &
47.18 &
61.73 &
62.05 &
64.03 &
69.99 & 88.5 \\[0.2em] 

\rowcolor{Blue2}
Verify &
53.40 &
58.80 &
47.02 &
65.78 &
67.00 &
71.45 &
75.45 & 90.1 \\[0.2em] 

Global &
24.70 &
20.19 &
8.64 &
27.22 &
56.57 &
60.29 &
60.82 & 92.3 \\[0.2em] 

Object &
49.96 &
54.00 &
47.33 &
74.33 &
75.90 &
78.45 &
81.49 & 88.1 \\[0.2em] 

Attribute &
34.89 &
42.67 &
22.66 &
48.28 &
50.91 &
53.88 &
59.82 & 90.7\\[0.2em] 

Relation &
22.88 &
20.16 &
11.60 &
33.24 &
39.45 &
42.84 &
46.16 & 89.2 \\[0.2em] 

Category &
15.26 &
17.31 &
3.56 &
22.33 &
37.49 &
41.18 &
44.38 & 90.3 \\[0.2em] 

\rowcolor{Blue2}
Distribution &
130.86 &
21.56 &
19.99 &
17.93 &
7.46 &
5.98 &
5.34 & - \\[0.2em]

Grounding &
- &
- &
- &
- &
- &
78.47 &
82.24 & - \\[0.2em]			

\rowcolor{Blue2}
Validity &
89.02 &
84.44 &
35.78 &
96.39 &
96.02 &
96.18 &
96.16 & 98.9 \\[0.2em] 

\rowcolor{Blue2}
Plausibility &
75.34 &
84.42 &
34.84 &
87.30 &
84.25 &
84.57 &
84.48 & 97.2 \\[0.2em] 

Consistency &
51.78 &
54.34 &
62.40 &
68.68 &
74.57 &
78.71 &
81.59 & 98.4 \\[0.2em] 

\rowcolor{Blue2}
\textbf{Accuracy} &
\textbf{28.93} &
\textbf{31.31} &
\textbf{17.82} &
\textbf{41.07} &
\textbf{46.55} &
\textbf{49.74} &
\textbf{54.06} & \textbf{89.3} \\[0.2em] 

\end{tabular}
\vspace*{2mm}
\caption{Results for baselines and state-of-the-art models on the GQA dataset. All results refer to the test set. Models are evaluated for overall accuracy as well as accuracy per type. In addition, they are evaluated by validity, plausibility, distribution, consistency, and when possible, grounding metrics. Please refer to the text for further detail.}
\label{table2}
\end{table*}


Apart from the standard accuracy metric and the more detailed type-based diagnosis our dataset supports, we introduce five new metrics to get further insight into visual reasoning methods and point to missing capabilities we believe coherent reasoning models should possess. 

\textbf{Consistency}. This metric measures responses consistency across different questions. Recall that in  \secref{exppp}, we used the questions' semantic representation to derive equivalence and entailment relations between them. When being presented with a new question, any learner striving to be trustworthy should not contradict its previous answers. It should not answer \textit{green} to a new question about an apple it has just identified as \textit{red}. 

For each question-answer pair $(q,a)$, we define a set $E_q={q_1, q_2, \dots, q_n}$ of entailed questions, the answers to which can be unambiguously inferred given $(q,a)$. For instance, given the question-answer pair \textit{Is there a red apple to the left of the white plate? yes}, we can infer the answers to questions such as \textit{Is the plate to the right of the apple?}, \textit{Is there a red fruit to the left of the plate?}, \textit{What is the white thing to the right of the apple?}, etc. For each question $q$ in $Q$ -- the set of questions \textit{the model answered correctly}, we measure the model's accuracy over the entailed questions $E_q$ and then average these scores across all questions in $Q$. 

We see that while people have exceptional consistency of 98.4\%, even the best models are inconsistent in about 1 out of 5 questions, and models such as LSTM contradict themselves almost half the time. Achieving high consistency may require deeper understanding of the question semantics in the context of the image, and, in contrast with accuracy, is more robust against  educated guesses as it inspects connections between related questions, and thus may serve as a better measure of models' true visual understanding skills.



\textbf{Validity} and \textbf{Plausibility}. The validity metric checks whether a given answer is in the question scope, \eg responding some color to a color question. The plausibility score goes a step further, measuring whether the answer is reasonable, or makes sense, given the question (\eg elephant usually do not eat, say, pizza). Specifically, we check whether the answer occurs at least once in relation with the question's subject, across the whole dataset. Thus, we consider e.g., \textit{red} and \textit{green} as plausible apple colors and, conversely, \textit{purple} as implausible.\footnote{While the plausibility metric may not be fully precise especially for infrequent objects due to potential data scarcity issues, it may provide a good sense of the general level of world knowledge the model has acquired.} The experiments show that models fail to respond with plausible or even valid answers at least 5--15\% of the times, indicating limited comprehension of some questions. Given that these properties are noticeable statistics of the dataset's conditional answer distribution, not even depending on the specific images, we would expect a sound method to achieve higher scores.

\textbf{Distribution}. To get further insight into the extent to which methods manage to model the conditional answer distribution, we define the distribution metric, which measures the overall match between the true answer distribution and the model predicted distribution, using Chi-Square statistic \cite{chisquare}. It allows us to see if the model predicts not only the most common answers but also the less frequent ones. Indeed, the experiments demonstrate that the leading SOTA models score lower than the baselines (for this metric, lower is better), indicating increased capacity in fitting more subtle trends of the dataset's distribution.

\textbf{Grounding}. For attention-based models, the grounding score checks whether the model attends to regions within the image that are relevant to the question. For each dataset instance, we define a pointer $r$ to the visual region which the question or answer refer to, and measure the model's visual attention (probability) over that region. This metric allows us to evaluate the degree to which the model grounds its reasoning in the image, rather than just making educated guesses based on question priors or world tendencies. 

Indeed, the models mostly attend to the relevant regions in the image, with grounding scores of about 80\%. To verify the reliability of the metric, we further perform experiments with spatial features instead of the object-informed ones used by BottomUp \cite{bottomup} and MAC \cite{mac}, which lead to a much lower 43\% score, demonstrating that object-based features provide models with better granularity for the task, allowing them to focus on more pertinent regions than with the coarser spatial features.

\section{Conclusion}
In this paper, we introduced the GQA dataset for real-world visual reasoning and compositional question answering. We described the dataset generation process, provided baseline experiments and defined new measures to get more insight into models' behavior and performance. We believe this benchmark can help drive VQA research in the right directions of deeper semantic understanding, sound reasoning, enhanced robustness and improved consistency. A potential avenue towards such goals may involve more intimate integration between visual knowledge extraction and question answering, two flourishing fields that oftentimes have been pursued independently. We strongly hope that GQA will motivate and support the development of more compositional, interpretable and cogent reasoning models, to advance research in scene understanding and visual question answering.

\section{Acknowledgments}
We wish to thank Justin Johnson for discussions about the early versions of this work, and Ross Girshick for his inspirational talk at the VQA workshop 2018. We thank Ranjay Krishna, Eric Cosatto, Alexandru Niculescu-Mizil and the anonymous reviewers for helpful suggestions and comments. Stanford University gratefully acknowledges Facebook Inc., Samsung Electronics Co., Ltd., and the Defense Advanced Research Projects Agency (DARPA) Communicating with Computers (CwC) program under ARO prime contract no. W911NF15-1-0462 for generously supporting this work.

{\small
\bibliographystyle{ieee}
\bibliography{egbib}
}
\vspace*{-12mm}

\clearpage
\section{Dataset Visualizations}
\thispagestyle{empty}

\label{sec:visuals}
\begin{center}
\centering
\begin{minipage}[b]{1.0\textwidth}
\centering
\vspace*{-3mm}
\begin{figure}[H]
\subfloat{\includegraphics[width=0.4\linewidth]{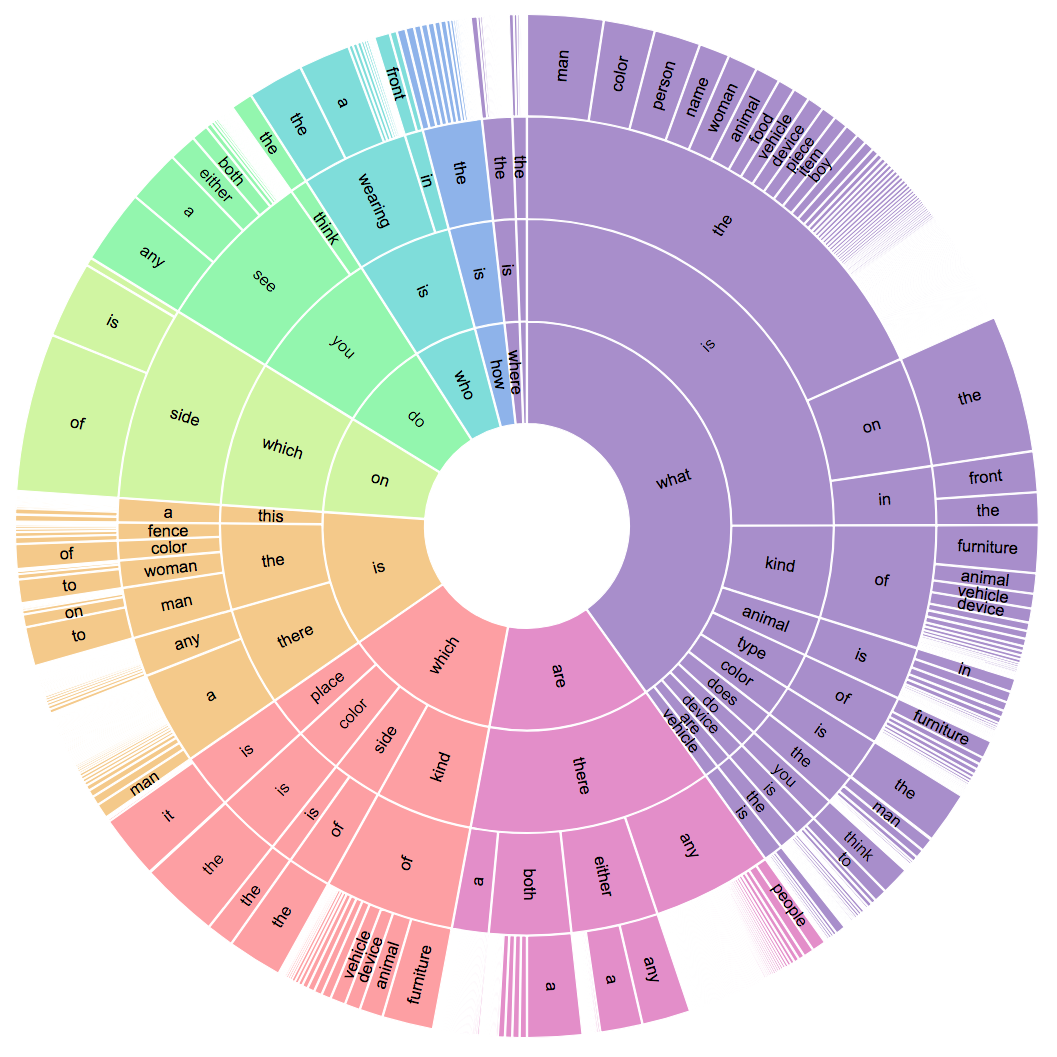}}
\hfill
\subfloat{\includegraphics[width=0.5\linewidth]{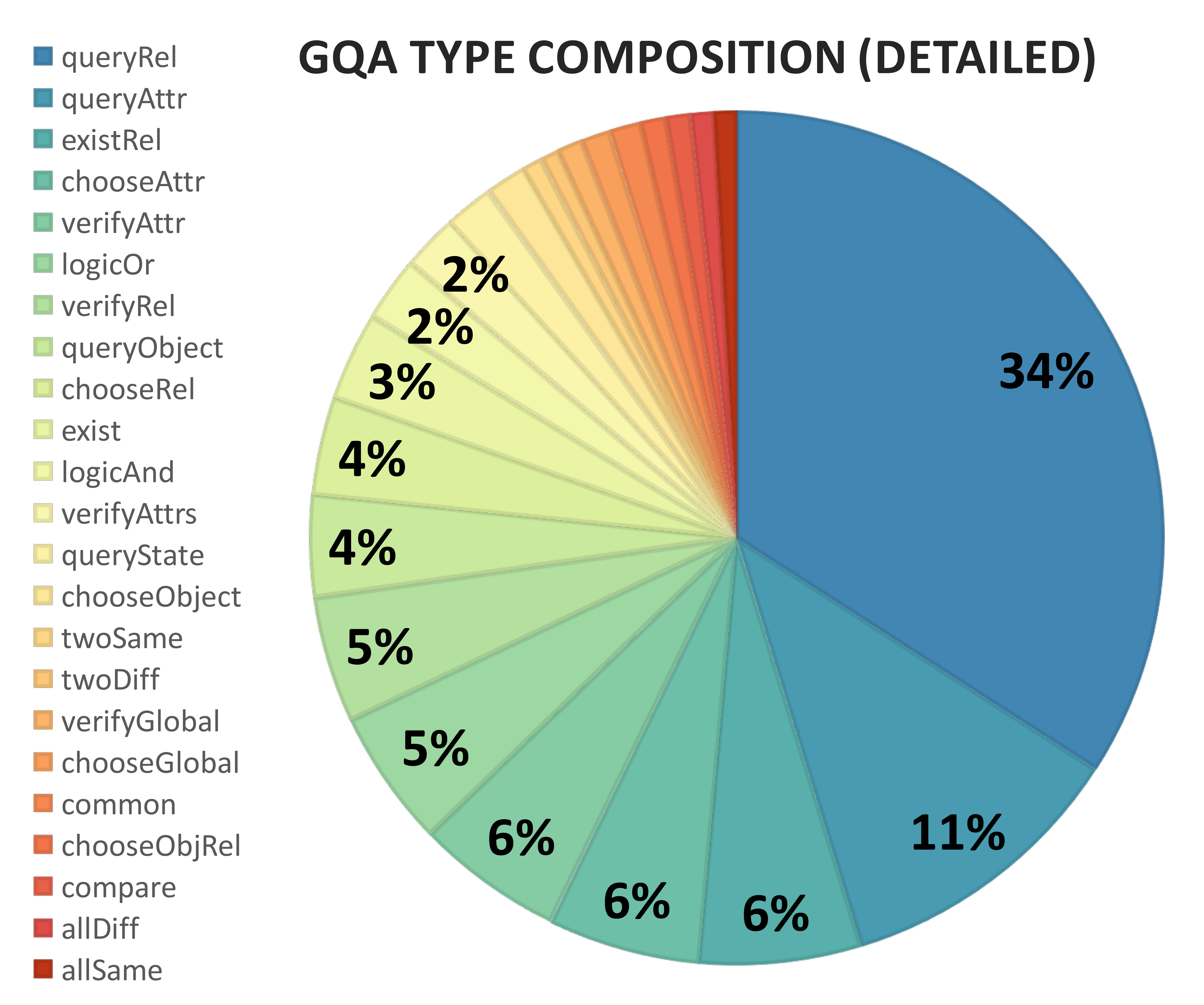}}
\end{figure}
\begin{figure}[H]
\subfloat{\includegraphics[width=0.24\linewidth]{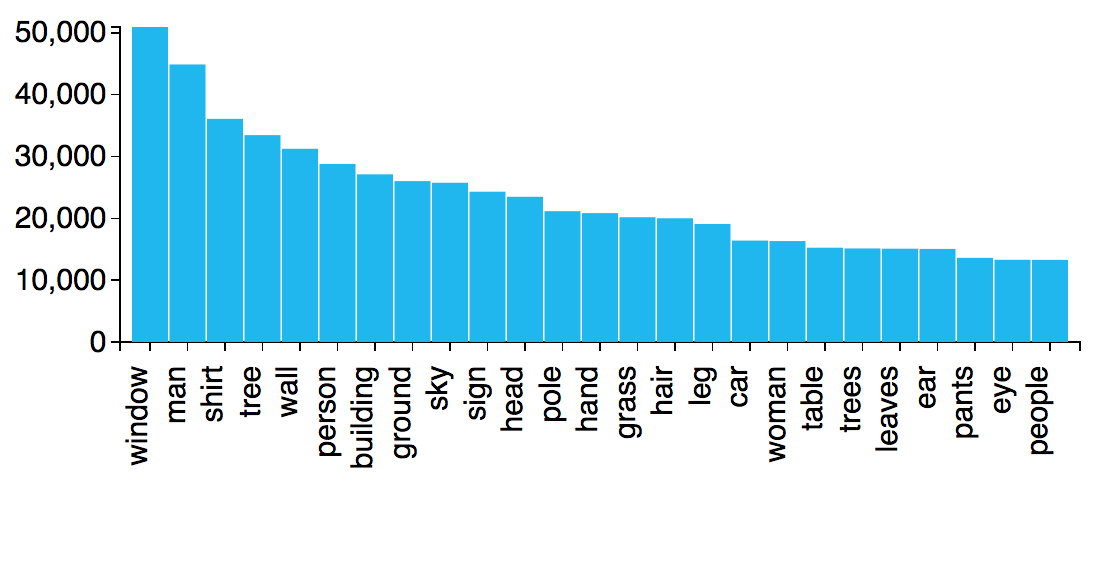}}
\hfill
\subfloat{\includegraphics[width=0.24\linewidth]{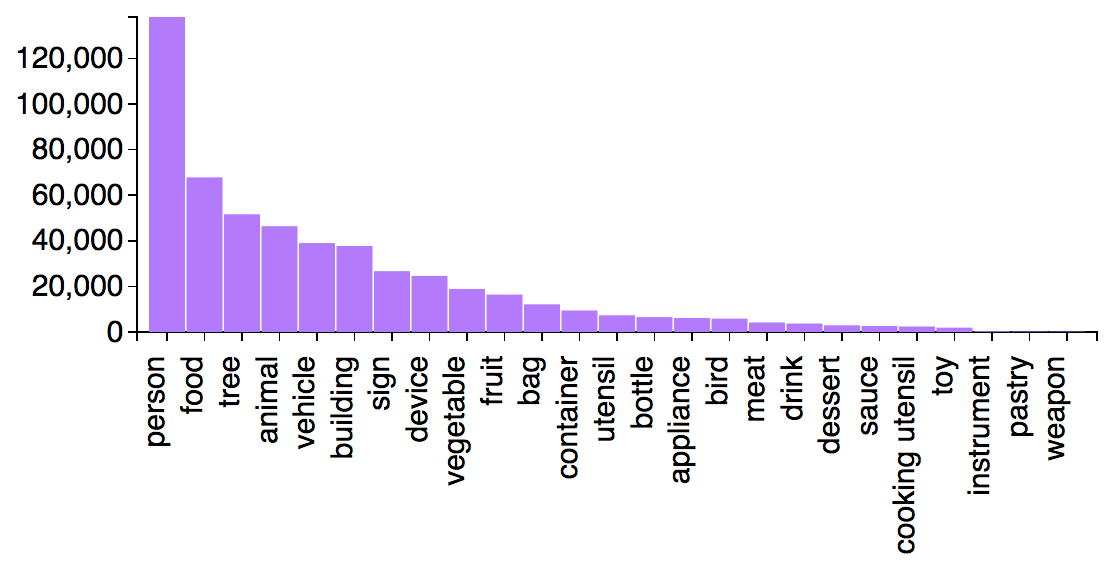}}
\hfill
\subfloat{\includegraphics[width=0.24\linewidth]{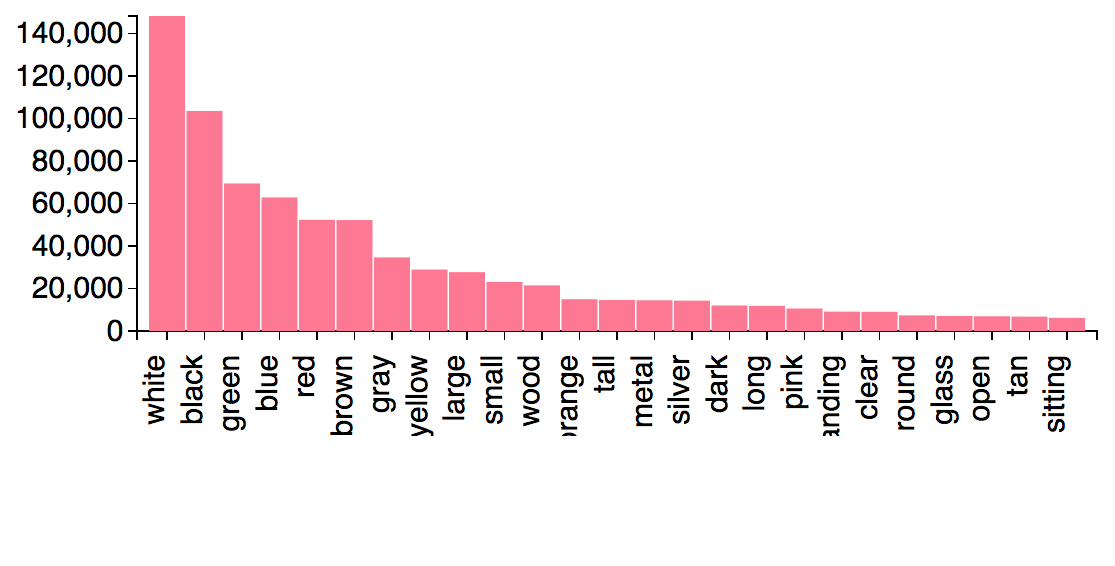}}
\hfill
\subfloat{\includegraphics[width=0.24\linewidth]{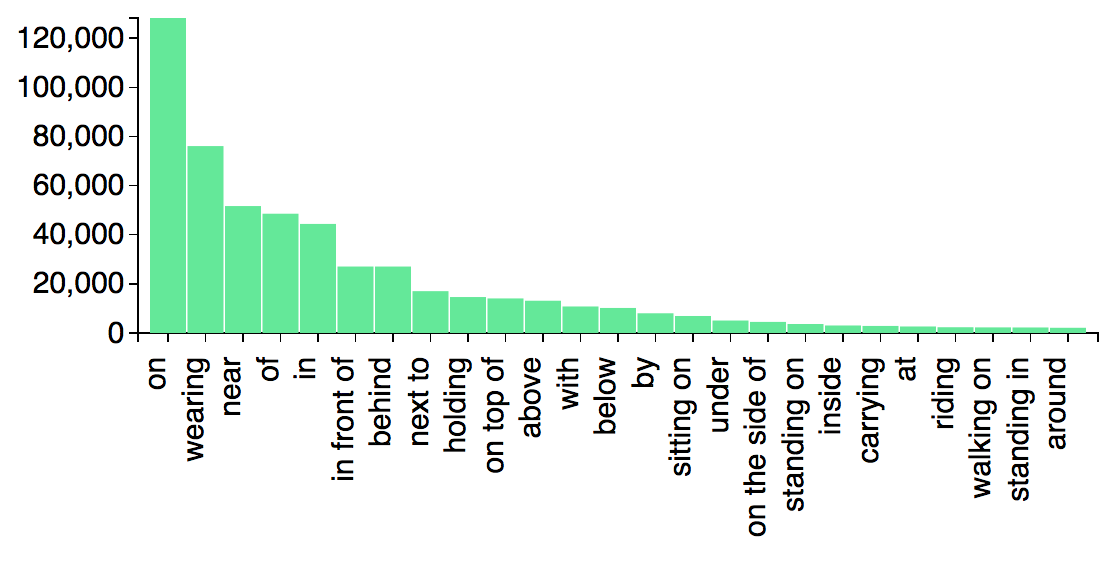}}
\end{figure}
\vspace*{-1.5mm}
\begin{figure}[H]
\subfloat{\includegraphics[width=0.29\linewidth]{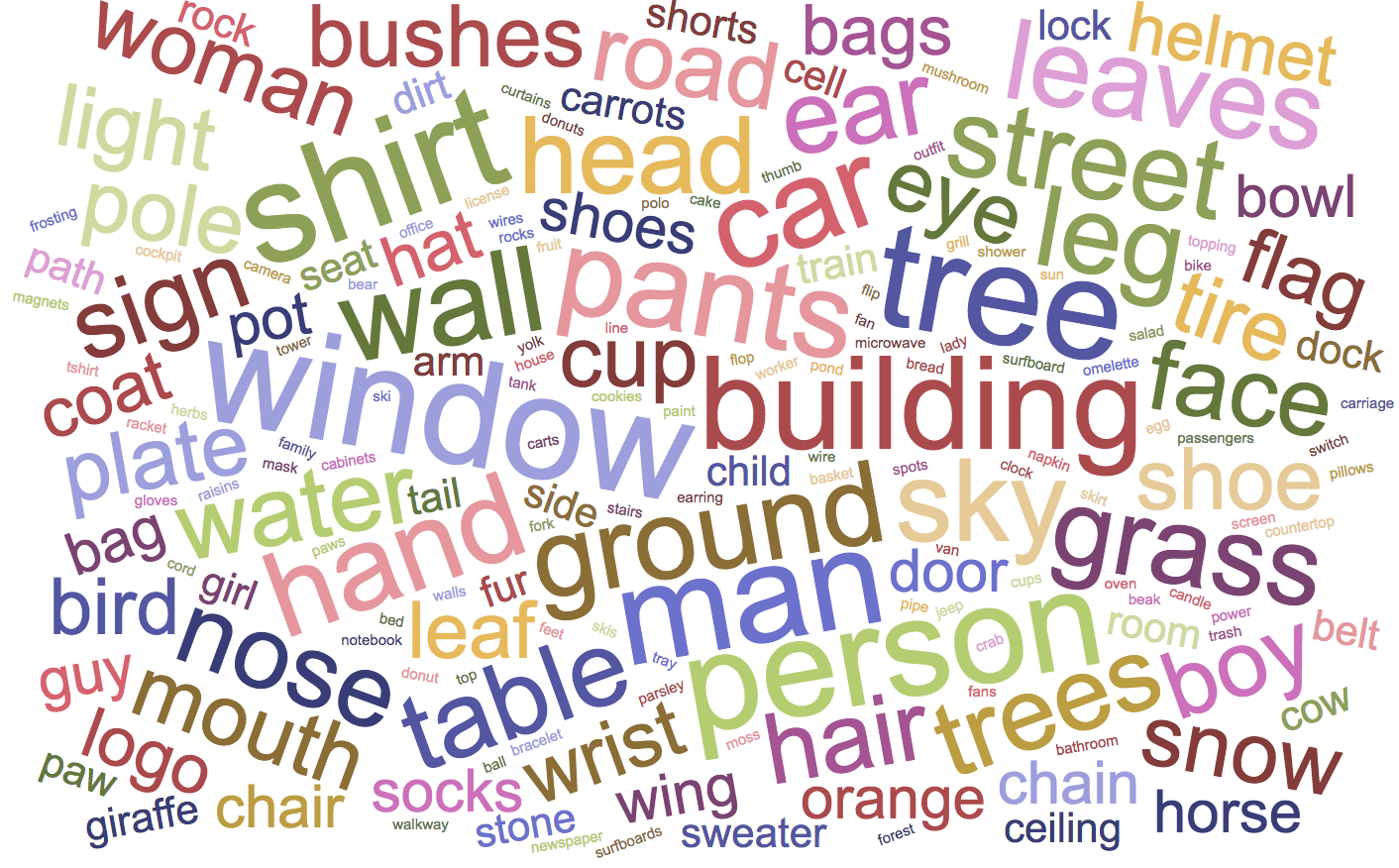}}
\hfill
\subfloat{\includegraphics[width=0.29\linewidth]{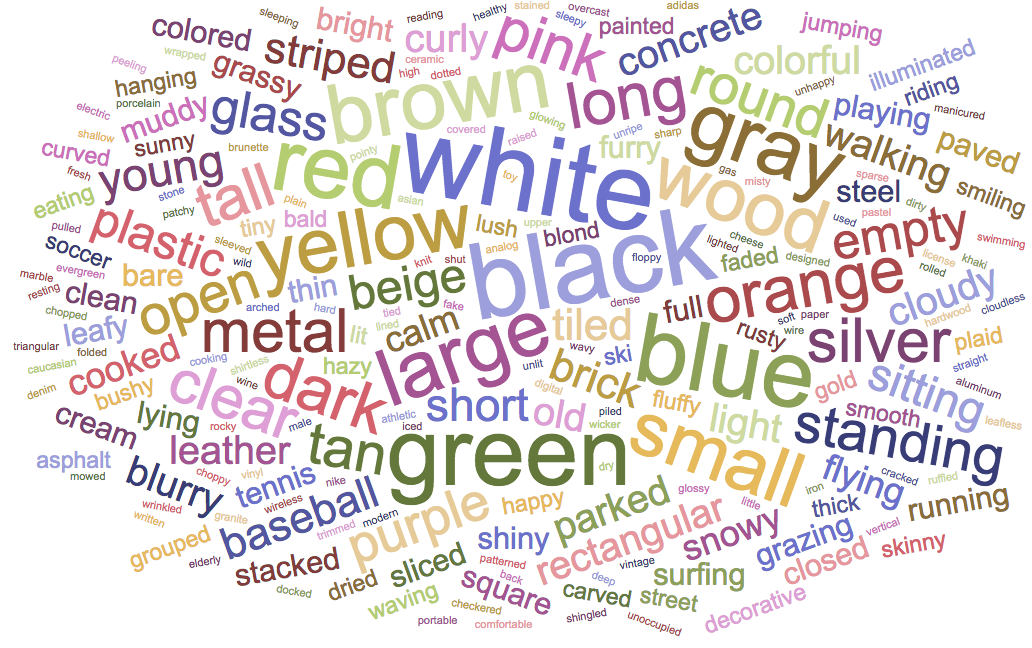}}
\hfill
\subfloat{\includegraphics[width=0.29\linewidth]{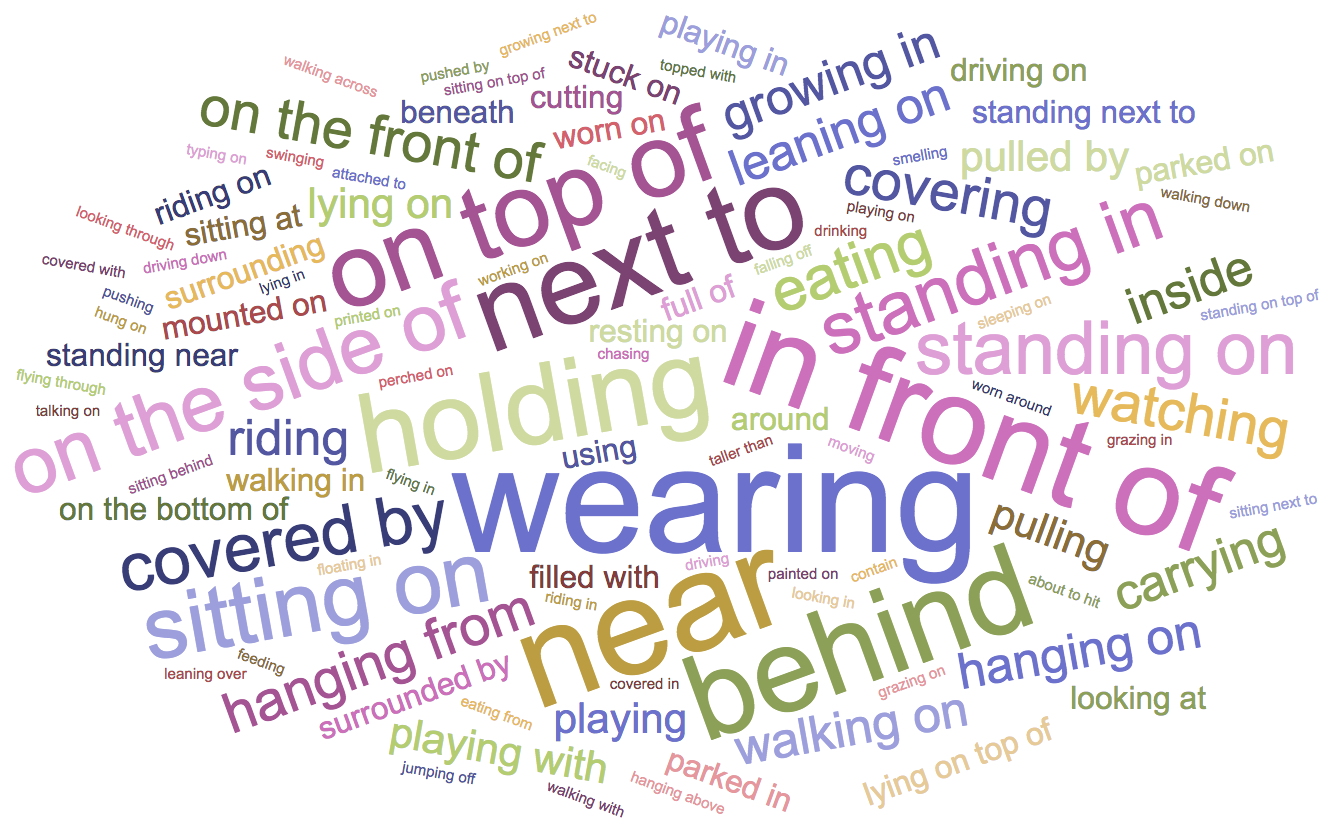}}
\vspace*{7.5mm}
\caption{\textbf{Top left}: Distribution of GQA questions by first four words. The arc length is proportional to the number of questions containing that prefix. White areas correspond to marginal contributions. \textbf{Top right}: question type distribution; please refer to the table \ref{table3} for details about each type. \textbf{Middle rows}: Occurrences number of the most frequent objects, categories, attributes and relations (excluding left/right). \textbf{Third row}: Word clouds for frequent objects, attributes and relations.} 
\end{figure}
\end{minipage}\qquad
\end{center}

\clearpage
\begin{center}
\centering
\begin{minipage}[b]{1.0\textwidth}
\centering
\vspace*{-11mm}
\begin{figure}[H]
\begin{minipage}{0.32\textwidth}

\subfloat{
\includegraphics[width=0.86\textwidth]{imgs/qdist.pdf}
}
\vspace*{-1mm}
\subfloat{
\includegraphics[width=0.84\textwidth]{imgs/scake1.pdf}
}
\centering
\vspace*{-2.5mm}
\subfloat{
\includegraphics[width=0.74\textwidth]{imgs/cakenew2.pdf}
}

\end{minipage}
\begin{minipage}{0.65\textwidth}

\subfloat{
\includegraphics[width=\textwidth]{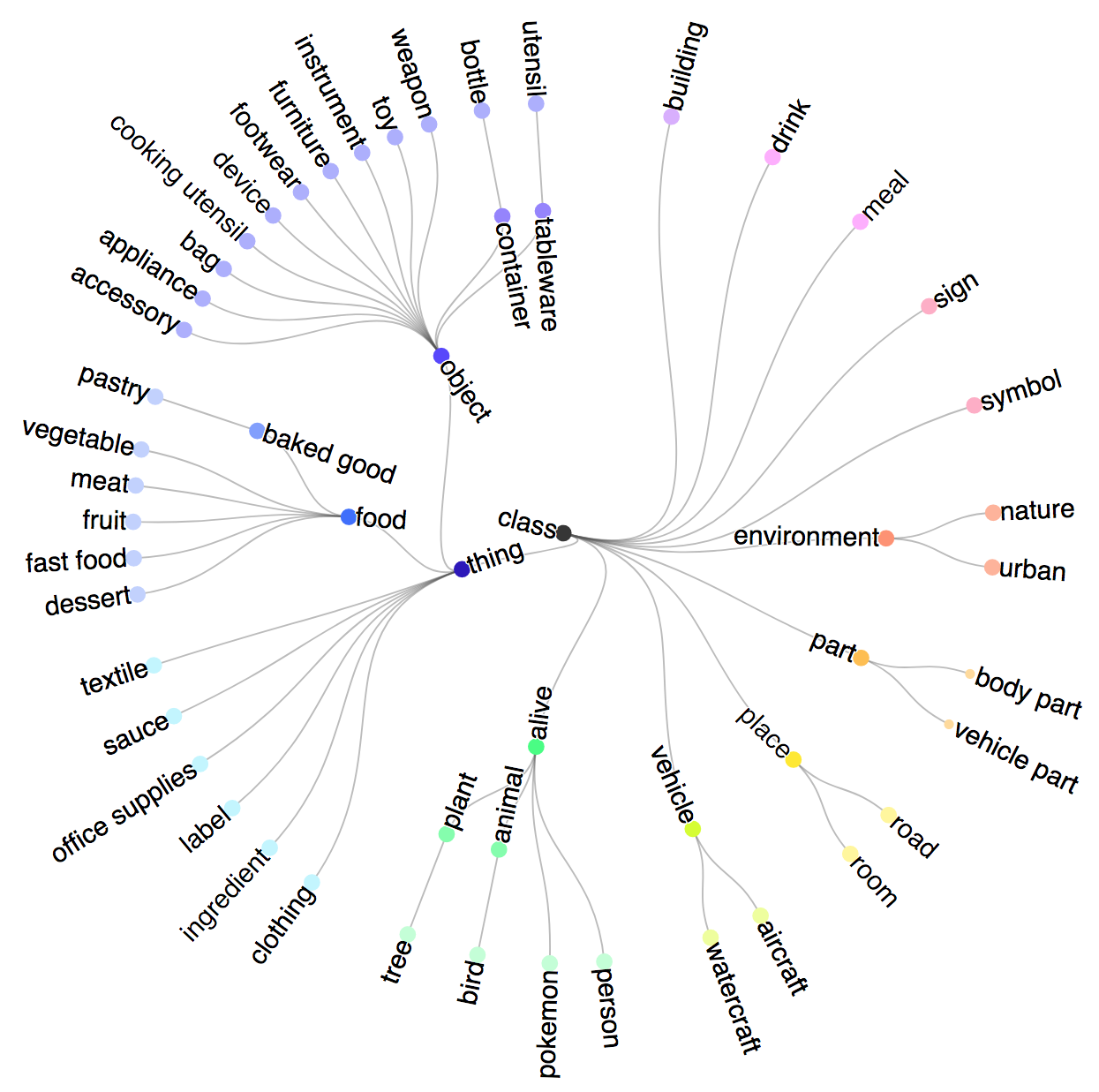}
}
\end{minipage}

\caption{\textbf{Top left}: Question length distribution for VQA datasets: we can see that GQA has a diverse
range of lengths compared to all other datasets except synthetic CLEVR. \textbf{Left}: GQA Question structural and semantic type distributions. \textbf{Right}: The object class hierarchy we have created as part of the dataset construction process.} 
\label{fig:ontology}
\end{figure}

\centering
\scriptsize 
\begin{tabular}{lcccll}
\rowcolor{Blue1}
\textbf{Type} & \textbf{Open/Binary} &\textbf{Semantic} & \textbf{Structural} & \textbf{Form} & \textbf{Example} \\ 

queryGlobal & open & query & global & select: \texttt{scene}/query: \texttt{type} & How is the weather in the image? \\[0.2em] 

verifyGlobal & binary & verify & global & select: \texttt{scene}/verify \texttt{type}: \texttt{attr} & Is it cloudy today? \\[0.2em] 

chooseGlobal & open & query & global & select: \texttt{scene}/choose \texttt{type}: \texttt{a}$|$\texttt{b} & Is it sunny or cloudy? \\[0.2em] 

\rowcolor{Blue2}
queryAttr & open & query & attribute & select: \texttt{obj}/\dots/query: \texttt{type} & What color is the apple? \\[0.2em] 

\rowcolor{Blue2}
verifyAttr & binary & verify & attribute & select: \texttt{obj}/\dots/verify \texttt{type}: \texttt{attr} & Is the apple red? \\[0.2em] 

\rowcolor{Blue2}
verifyAttrs & binary & logical & attribute & select: \texttt{obj}/\dots/verify \texttt{t1}: \texttt{a1}/verify \texttt{t2}: \texttt{a2}/and & Is the apple red and shiny? \\[0.2em] 

\rowcolor{Blue2}
chooseAttr & open & choose & attribute & select: \texttt{obj}/\dots/choose \texttt{type}: \texttt{a}$|$\texttt{b} & Is the apple green or red? \\[0.2em] 

exist & binary & verify & object & select: \texttt{obj}/\dots/exist & Is there an apple in the picture? \\[0.2em] 

existRel & binary & verify & relation & select: \texttt{subj}/\dots/relate (\texttt{rel}): \texttt{obj}/exist & Is there an apple on the black table? \\[0.2em] 

\rowcolor{Blue2}
logicOr & binary & logical & object & select: \texttt{obj1}/\dots/exist/select: \texttt{obj2}/\dots/exist/or & Do you see either an apple or a banana there? \\[0.2em] 

\rowcolor{Blue2}
logicAnd & binary & logical & obj/attr & select: \texttt{obj1}/\dots/exist/select: \texttt{obj2}/\dots/exist/and & Do you see both green apples and bananas there? \\[0.2em] 

queryObject & open & query & category & select: \texttt{category}/\dots/query: name & What kind of fruit is on the table? \\[0.2em] 

chooseObject & open & choose & category & select: \texttt{category}/\dots/choose: \texttt{a}$|$\texttt{b} & What kind of fruit is it, an apple or a banana? \\[0.2em] 

\rowcolor{Blue2}
queryRel & open & query & relation & select: \texttt{subj}/\dots/relate (\texttt{rel}): \texttt{obj}/query: name & What is the small girl wearing? \\[0.2em] 

\rowcolor{Blue2}
verifyRel & binary & verify & relation & select: \texttt{subj}/\dots/verifyRel (\texttt{rel}): \texttt{obj} & Is she wearing a blue dress? \\[0.2em] 

\rowcolor{Blue2}
chooseRel & open & choose & relation & select: \texttt{subj}/\dots/chooseRel (\texttt{r1}$|$\texttt{r2}): \texttt{obj} & Is the cat to the left or to the right of the flower? \\[0.2em] 

\rowcolor{Blue2}
chooseObjRel & open & choose & relation & select: \texttt{subj}/\dots/relate (\texttt{rel}): \texttt{obj}/choose: \texttt{a}$|$\texttt{b} & What is the boy eating, an apple or a slice of pizza? \\[0.2em] 

compare & binary & compare & object & select: \texttt{obj1}/\dots/select: \texttt{obj2}/\dots/compare \texttt{type} & Who is taller, the boy or the girl? \\[0.2em] 

common & open & compare & object & select: \texttt{obj1}/\dots/select: \texttt{obj2}/\dots/common & What is common to the shirt and the flower? \\[0.2em] 

twoSame & verify & compare & object & select: \texttt{obj1}/\dots/select: \texttt{obj2}/\dots/same & Does the shirt and the flower have the same color? \\[0.2em] 

twoDiff & verify & compare & object & select: \texttt{obj1}/\dots/select: \texttt{obj2}/\dots/different & Are the table and the chair made of different materials? \\[0.2em] 

allSame & verify & compare & object & select: \texttt{allObjs}/same & Are all the people there the same gender? \\[0.2em] 

allDiff & verify & compare & object & select: \texttt{allObjs}/different & Are the animals in the image of different types? \\[0.2em]

\end{tabular}
\newline
\captionof{table}{Functions Catalog for all the GQA question types. For each question we mention its structural and semantic types (refer to table \ref{table3} for further details), a functional program template and a typical example of a generated question.}
\label{table3}

\end{minipage}\qquad

\end{center}

\clearpage

\begin{center}
\footnotesize
\centering
\begin{minipage}{1.0\textwidth}
\begin{minipage}{0.205\textwidth}
\includegraphics[width=\linewidth]{imgs/2416306.jpg}
\end{minipage}\qquad
\begin{minipage}{0.42\textwidth}

\begin{flushleft}
\textbf{{\color{blue} GQA}} \\
\textbf{1}. What is the \textbf{{\color{violet} woman}} to the right of the \textbf{{\color{teal} boat}} holding? umbrella \\ 
\textbf{2}. Are there \textbf{{\color{olive} men}} to the left of the \textbf{{\color{violet} person}} that is holding the \textbf{{\color{orange} umbrella}}? no \\ 
\textbf{3}. What color is the \textbf{{\color{orange} umbrella}} the \textbf{{\color{violet} woman}} is holding? purple
\end{flushleft}
\end{minipage}\qquad
\begin{minipage}{0.32\textwidth}

\textbf{{\color{red} VQA}} \\
\textbf{1}. Why is the person using an umbrella? \\
\textbf{2}. Is the picture edited? \\
\textbf{3}. What's the color of the umbrella? \\

\end{minipage}\qquad
\end{minipage}\qquad
\vfill

\begin{minipage}{1.0\textwidth}
\begin{minipage}{0.205\textwidth}
\includegraphics[width=\linewidth]{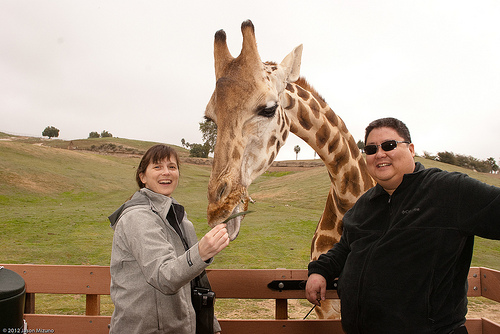}
\end{minipage}\qquad
\begin{minipage}{0.42\textwidth}

\textbf{{\color{blue} GQA}}

\textbf{1}. Is that a \textbf{{\color{teal} giraffe}} or an \textbf{{\color{purple} elephant}}? giraffe\\
\textbf{2}. Who is feeding the \textbf{{\color{teal} giraffe}} behind the \textbf{{\color{violet} man}}? lady \\
\textbf{3}. Is there any \textbf{{\color{orange} fence}} near the \textbf{{\color{teal} animal}} behind the \textbf{{\color{violet} man}}? yes \\
\textbf{4}. On which side of the image is the \textbf{{\color{violet} man}}? right \\
\textbf{5}. Is the \textbf{{\color{teal} giraffe}} is behind the \textbf{{\color{violet} man}}? yes \\

\end{minipage}\qquad
\begin{minipage}{0.32\textwidth}

\textbf{{\color{red} VQA}} \\
\textbf{1}. What animal is the lady feeding? \\
\textbf{2}. Is it raining? \\ 
\textbf{3}. Is the man wearing sunglasses?

\end{minipage}\qquad
\end{minipage}\qquad
\vfill

\begin{minipage}{1.0\textwidth}
\begin{minipage}{0.205\textwidth}
\includegraphics[width=\linewidth]{imgs/2417926.jpg}
\end{minipage}\qquad
\begin{minipage}{0.42\textwidth}

\textbf{{\color{blue} GQA}}

\textbf{1}. Is the \textbf{{\color{teal} person}}'s \textbf{{\color{orange} hair}} brown and long? yes \\
\textbf{2}. What \textbf{{\color{violet} appliance}} is to the left of the \textbf{{\color{teal} man}}? refrigerator \\
\textbf{3}. Is the \textbf{{\color{teal} man}} to the left or to the right of a \textbf{{\color{violet} refrigerator}}? right \\
\textbf{4}. Who is in front of the \textbf{{\color{violet} appliance}} on the left? man \\
\textbf{5}. Is there a \textbf{{\color{magenta} necktie}} in the picture that is not red? yes \\
\textbf{6}. What is the \textbf{{\color{teal} person}} in front of the \textbf{{\color{violet} refrigerator}} wearing? suit \\
\textbf{7}. What is hanging on the \textbf{{\color{purple} wall}}? picture \\
\textbf{8}. Does the \textbf{{\color{olive} vest}} have different color than the \textbf{{\color{magenta} tie}}? no \\
\textbf{9}. What is the color of the \textbf{{\color{brown} shirt}}? white \\
\textbf{10}. Is the color of the \textbf{{\color{olive} vest}} different than the \textbf{{\color{brown} shirt}}? yes \\
\end{minipage}\qquad
\begin{minipage}{0.32\textwidth}

\begin{flushleft}
\textbf{{\color{red} VQA}} \\
\textbf{1}. Does this man need a haircut? \\
\textbf{2}. What color is the guys tie? \\
\textbf{3}. What is different about the man's suit that shows this is for a special occasion? \\
\end{flushleft}

\end{minipage}\qquad
\end{minipage}\qquad
\vfill

\begin{minipage}{1.0\textwidth}
\begin{minipage}{0.205\textwidth}
\includegraphics[width=\linewidth]{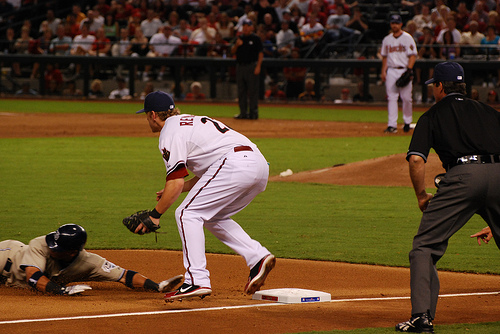}
\end{minipage}\qquad
\begin{minipage}{0.42\textwidth}

\textbf{{\color{blue} GQA}} \\
\textbf{1}. Who wears the \textbf{{\color{magenta} gloves}}? player \\
\textbf{2}. Are there any \textbf{{\color{orange} horses}} to the left of the \textbf{{\color{teal} man}}? no \\
\textbf{3}. Is the \textbf{{\color{teal} man}} to the right of the \textbf{{\color{violet} player}} that wears gloves? no \\
\textbf{4}. Is there a \textbf{{\color{purple} bag}} in the picture? no \\
\textbf{5}. Do the \textbf{{\color{olive}hat}} and the \textbf{{\color{brown}plate}} have different colors? yes \\

\end{minipage}\qquad
\begin{minipage}{0.32\textwidth}

\textbf{{\color{red} VQA}}

\textbf{1}. What is the man holding? \\
\textbf{2}. Where are the people playing? \\
\textbf{3}. Is the player safe? \\
\textbf{4}. What is the sport being played? \\

\end{minipage}\qquad
\end{minipage}\qquad

\vfill

\begin{minipage}{1.0\textwidth}
\begin{minipage}{0.205\textwidth}
\includegraphics[width=\linewidth]{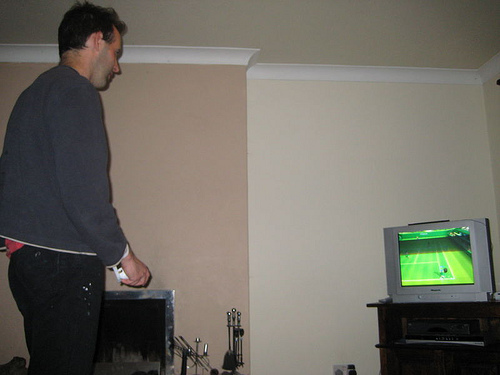}
\end{minipage}\qquad
\begin{minipage}{0.42\textwidth}
\begin{flushleft}
\textbf{{\color{blue} GQA}} \\
\textbf{1}. What is the \textbf{{\color{violet}person}} doing? playing \\
\textbf{2}. Is the \textbf{{\color{teal}entertainment center}} at the bottom or at the top? bottom \\
\textbf{3}. Is the \textbf{{\color{teal}entertainment center}} wooden and small? yes \\
\textbf{4}. Are the pants blue? no \\
\textbf{5}. Do you think the \textbf{{\color{magenta}controller}} is red? no \\
\end{flushleft}
\end{minipage}\qquad
\begin{minipage}{0.32\textwidth}

\textbf{{\color{red} VQA}}

\textbf{1}. What colors are the walls? \\
\textbf{2}. What game is the man playing? \\
\textbf{3}. Why do they stand to play? \\

\end{minipage}\qquad
\end{minipage}\qquad

\vfill

\begin{minipage}{1.0\textwidth}
\begin{minipage}{0.205\textwidth}
\includegraphics[width=\linewidth]{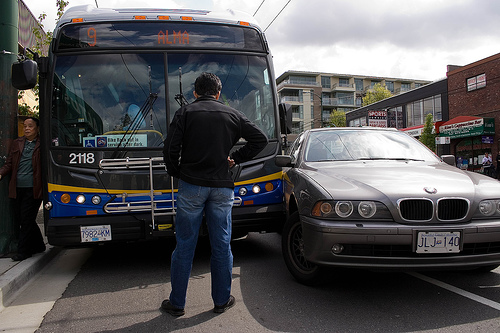}
\end{minipage}\qquad
\begin{minipage}{0.42\textwidth}

\textbf{{\color{blue} GQA}}

\textbf{1}. Are there any \textbf{{\color{teal}coats}}? yes \\
\textbf{2}. Do you see a red \textbf{{\color{teal}coat}} in the image? no \\
\textbf{3}. Is the \textbf{{\color{orange}person}} that is to the left of the \textbf{{\color{violet}man}} exiting a \textbf{{\color{olive}truck}}? no \\
\textbf{4}. Which place is this? road \\

\end{minipage}\qquad
\begin{minipage}{0.32\textwidth}

\textbf{{\color{red} VQA}}

\textbf{1}. Where is the bus driver? \\
\textbf{2}. Why is the man in front of the bus? \\
\textbf{3}. What numbers are repeated in the bus number? \\

\end{minipage}\qquad
\end{minipage}\qquad
\vfill

\begin{minipage}{1.0\textwidth}
\begin{minipage}{0.205\textwidth}
\includegraphics[width=\linewidth]{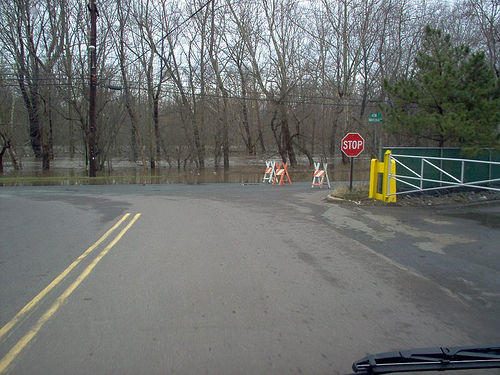}
\end{minipage}\qquad
\begin{minipage}{0.42\textwidth}

\textbf{{\color{blue} GQA}}

\textbf{1}. What is in front of the green \textbf{{\color{magenta}fence}}? gate \\
\textbf{2}. Of which color is the \textbf{{\color{teal}gate}}? silver \\
\textbf{3}. Where is this? street  \\
\textbf{4}. What color is the \textbf{{\color{magenta}fence}} behind the \textbf{{\color{teal}gate}}? green  \\
\textbf{5}. Is the \textbf{{\color{magenta}fence}} behind the \textbf{{\color{teal}gate}} both brown and metallic? no  \\

\end{minipage}\qquad
\begin{minipage}{0.32\textwidth}

\textbf{{\color{red} VQA}}

\textbf{1}. What are the yellow lines called? \\
\textbf{2}. Why don't the trees have leaves? \\
\textbf{3}. Where is the stop sign? \\

\end{minipage}\qquad
\captionof{figure}{Examples of questions from GQA and VQA, for the same  images. As the examples demonstrate, GQA questions tend to involve more elements from the image compared to VQA questions, and are longer and more compositional as well. Conversely, VQA questions tend to be a bit more ambiguous and subjective, at times with no clear and conclusive answer. Finally, we can see that GQA provides more questions for each image and thus covers it more thoroughly than VQA.}
\label{fig:examples2}
\end{minipage}\qquad
\end{center}


\clearpage
\section{Dataset Balancing}
\label{sec:balancing2}

\vspace*{-11mm}
\begin{center}
\centering
\begin{minipage}[b]{1.0\textwidth}
\centering
\begin{figure}[H]
\subfloat{\includegraphics[width=0.5\linewidth]{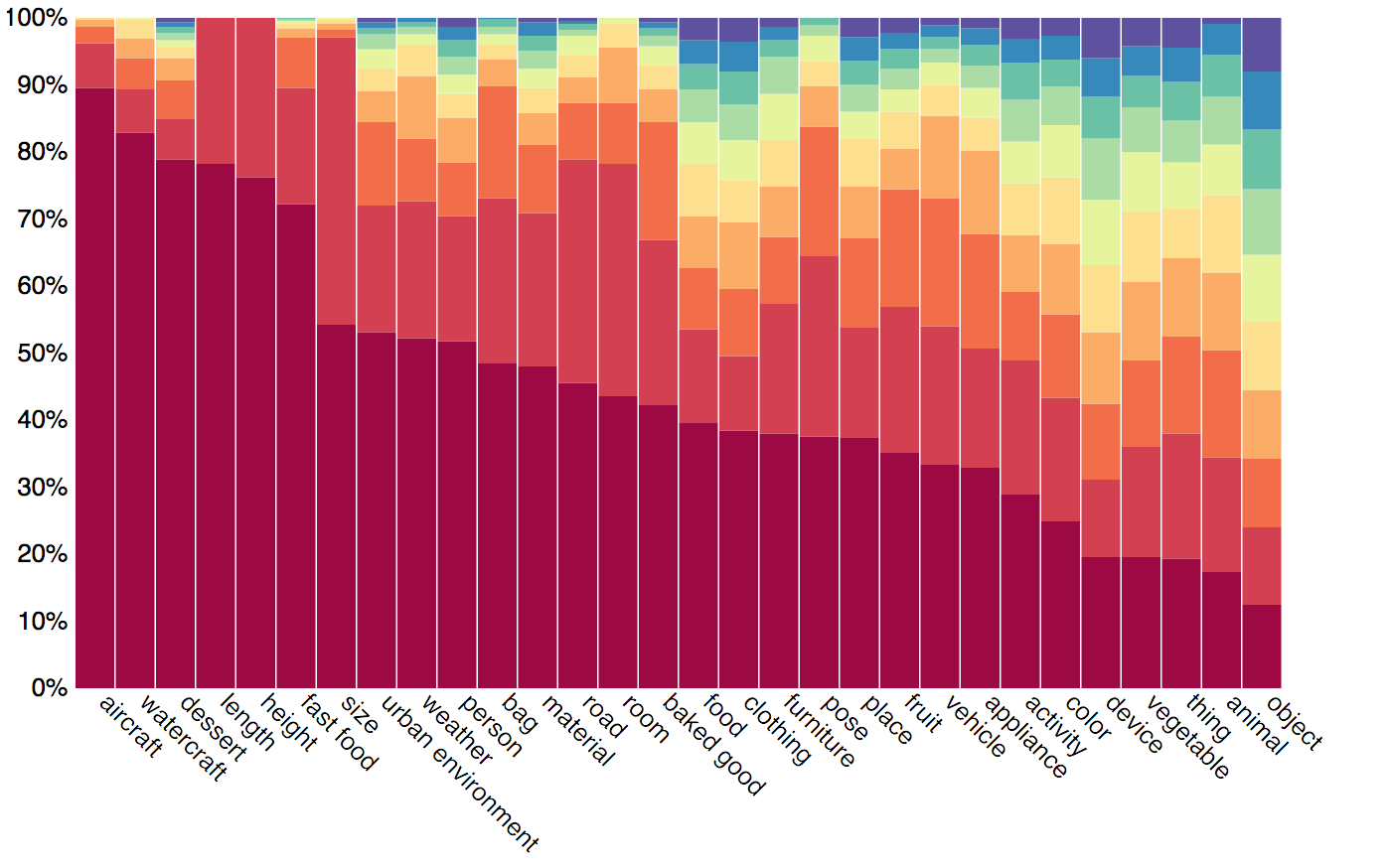}}
\hfill
\subfloat{\includegraphics[width=0.5\linewidth]{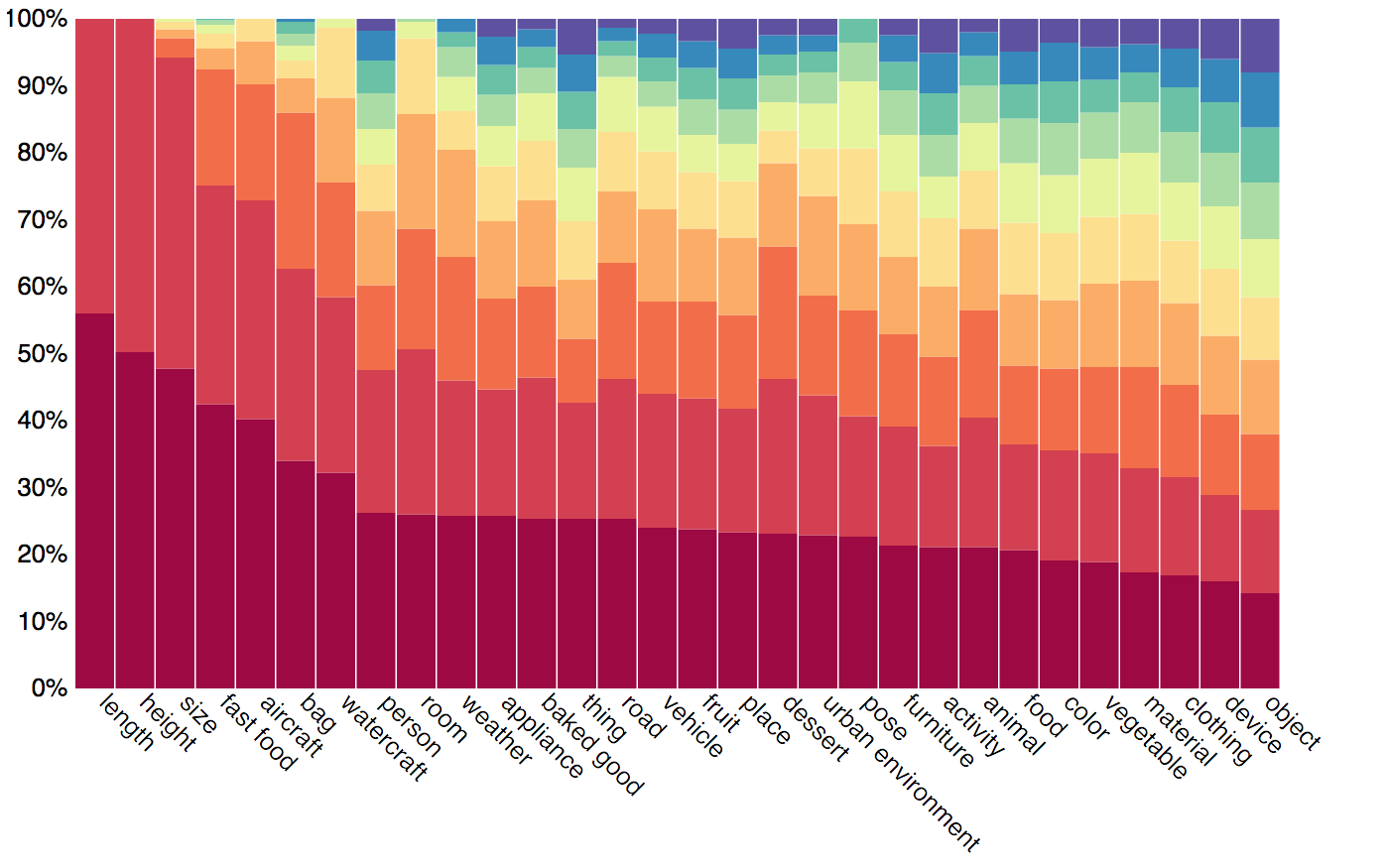}}

\vspace*{-2mm}
\subfloat{\includegraphics[width=0.5\linewidth]{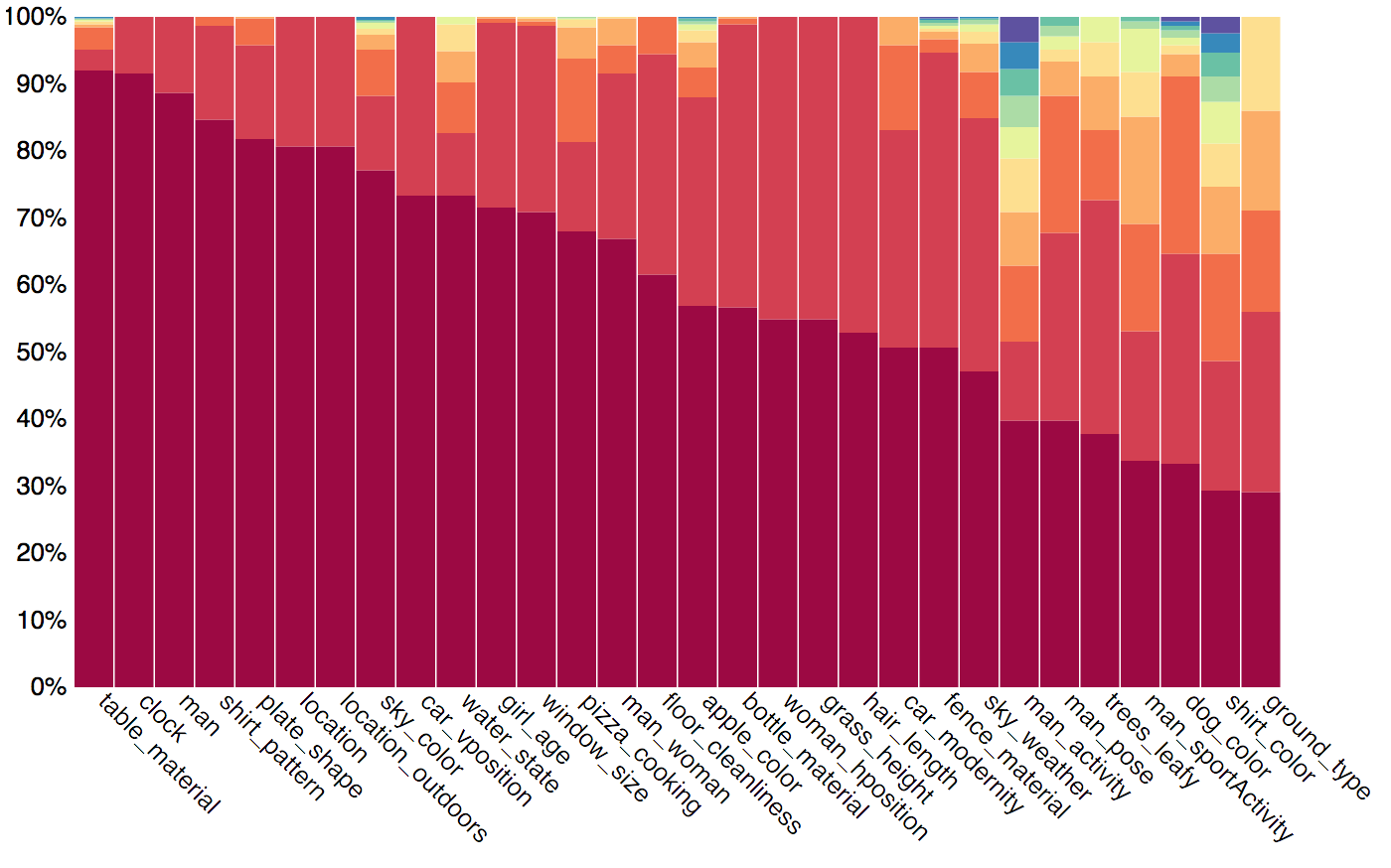}}
\hfill
\subfloat{\includegraphics[width=0.5\linewidth]{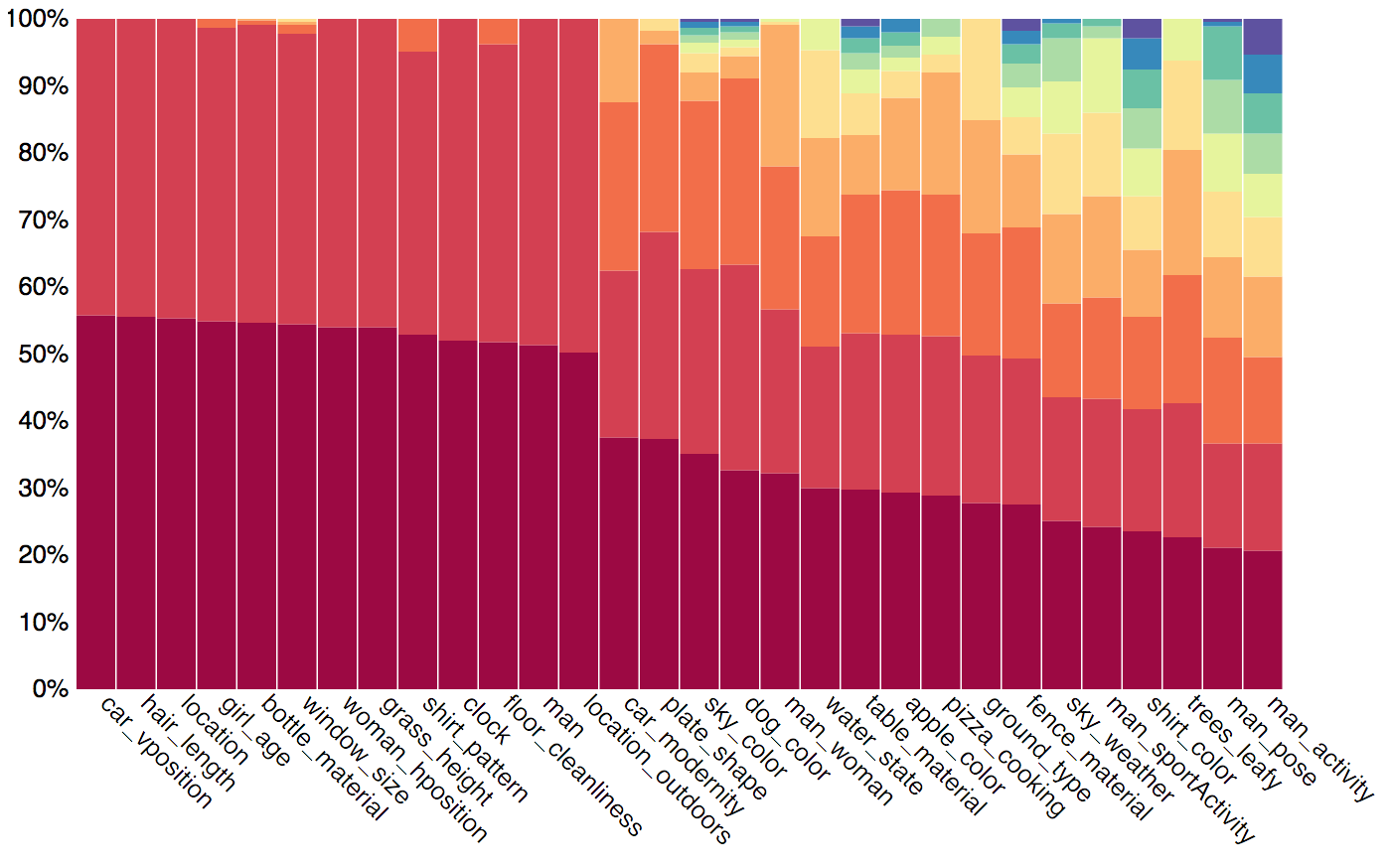}}
\caption{Impact of the dataset balancing on the conditional answer distribution: The left side shows the distribution before any balancing. We show the top 10 answers for a selection of question groups, where the column height corresponds to the relative frequency of each answer. The top row shows global question groups such as color questions, questions about animals, etc. while the bottom row shows local ones \eg \textit{apple-color}, \textit{table-material} etc (section 3.3, main paper). Indeed, we can see that the distributions are heavily biased. The right side shows the distributions after balancing, more uniform and with heavier tails, while intentionally retaining the original real-world tendencies up to a tunable degree.} 
\label{fig:balancing2}
\end{figure}
\end{minipage}\qquad


\vspace*{5mm}
\begin{minipage}[b]{1.0\textwidth}

\begin{multicols}{2}

As discussed in section 3.4 (main paper), given the original 22M auto-generated questions, we have performed answer-distribution balancing, similarities reduction and type-based sampling, reducing its size to a 1.7M balanaced dataset. The balancing is performed in an iterative manner: as explained in section 3.3, for each question group (\eg color questions), we iterate over the answer distribution, from the most to least frequent answers: $(a_i,c_i)$ when $a_i$ is the answer and $c_i$ is its count. In each iteration $i$, we downsample the head distribution ($a_j, j \leq i$) such that the ratio between the head and its complementary tail $\frac{\sum_{j \leq i} c_j}{1 - \sum_{j \leq i} c_j}$ will be bounded by $b$. While doing so, we also make sure to set minimum and maximum bounds on the frequency ratio $\frac{c_{i+1}}{c_i}$ of each pair of consequent answers $a_i,a_{i+1}$. The results of this process is shown in \figref{balancing2}. Indeed we can see how the distribution is ``pushed" away from the head and spreads over the tail, while intentionally maintaining the original real-world tendencies presented in the data, to retain its authenticity.

\section{Baselines Implementation Details}
\label{sec:baselines2}
In section 4.2 (main paper), we perform experiments over multiple baselines and state-of-the-art models. All CNN models use spatial features pre-trained on ImageNet \cite{Imagenet}, whereas state-of-the-art approaches such as bottomUp \cite{bottomup} and MAC \cite{mac} are based on object-based features produced by faster R-CNN detector \cite{frcnn}. All models use GloVe word embeddings of dimension 300 \cite{glove}. To allow a fair comparison, all the models use the same LSTM, CNN and classifier components, and so the only difference between the models stem from their core architectural design. 

\end{multicols}

\end{minipage}
\end{center}

\clearpage

\vspace*{-11mm}
\begin{center}
\centering
\begin{minipage}[b]{1.0\textwidth}
\centering
\begin{figure}[H]
\centering
\subfloat{\includegraphics[width=0.25\linewidth]{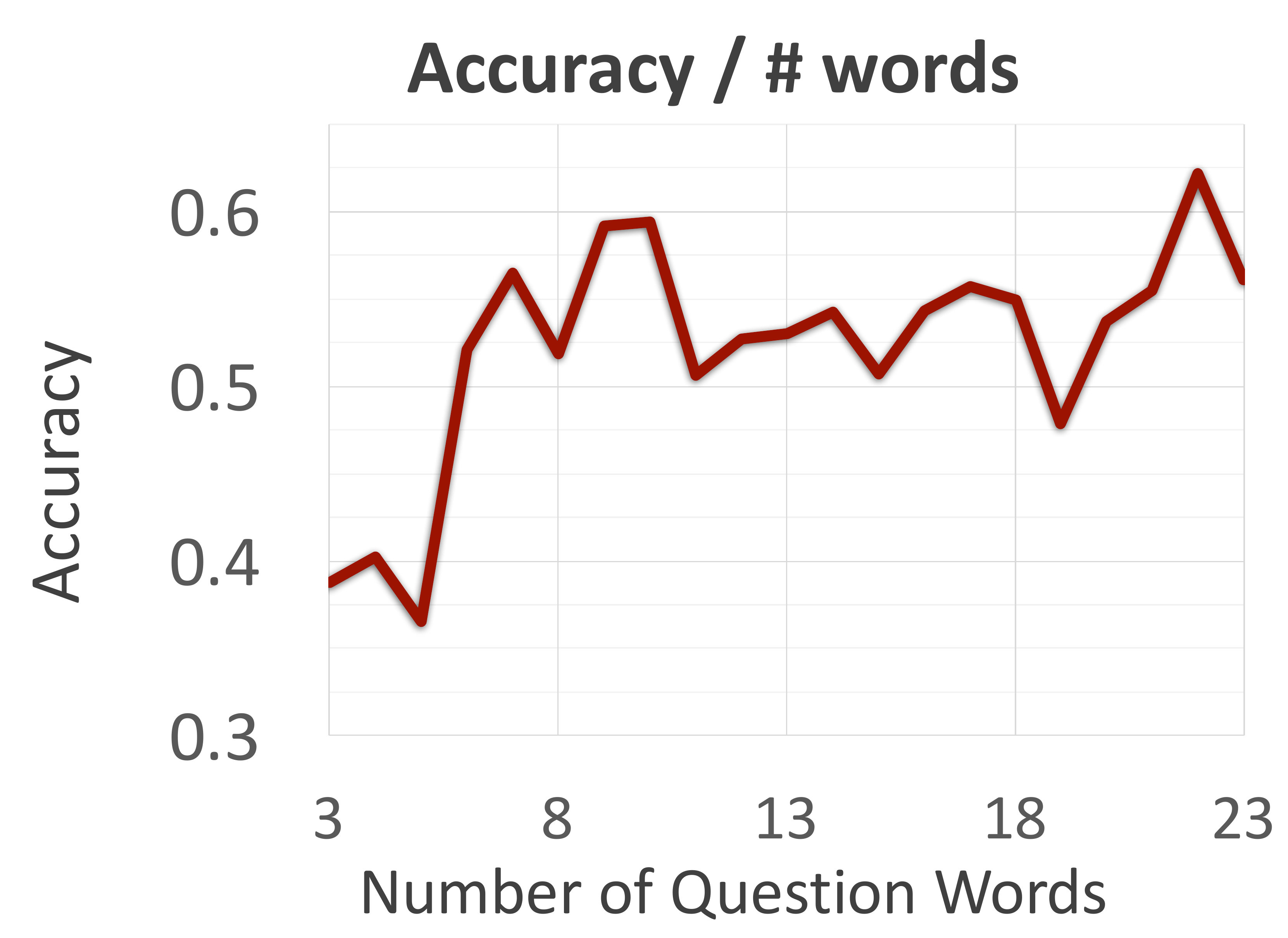}}
\hfill
\subfloat{\includegraphics[width=0.25\linewidth]{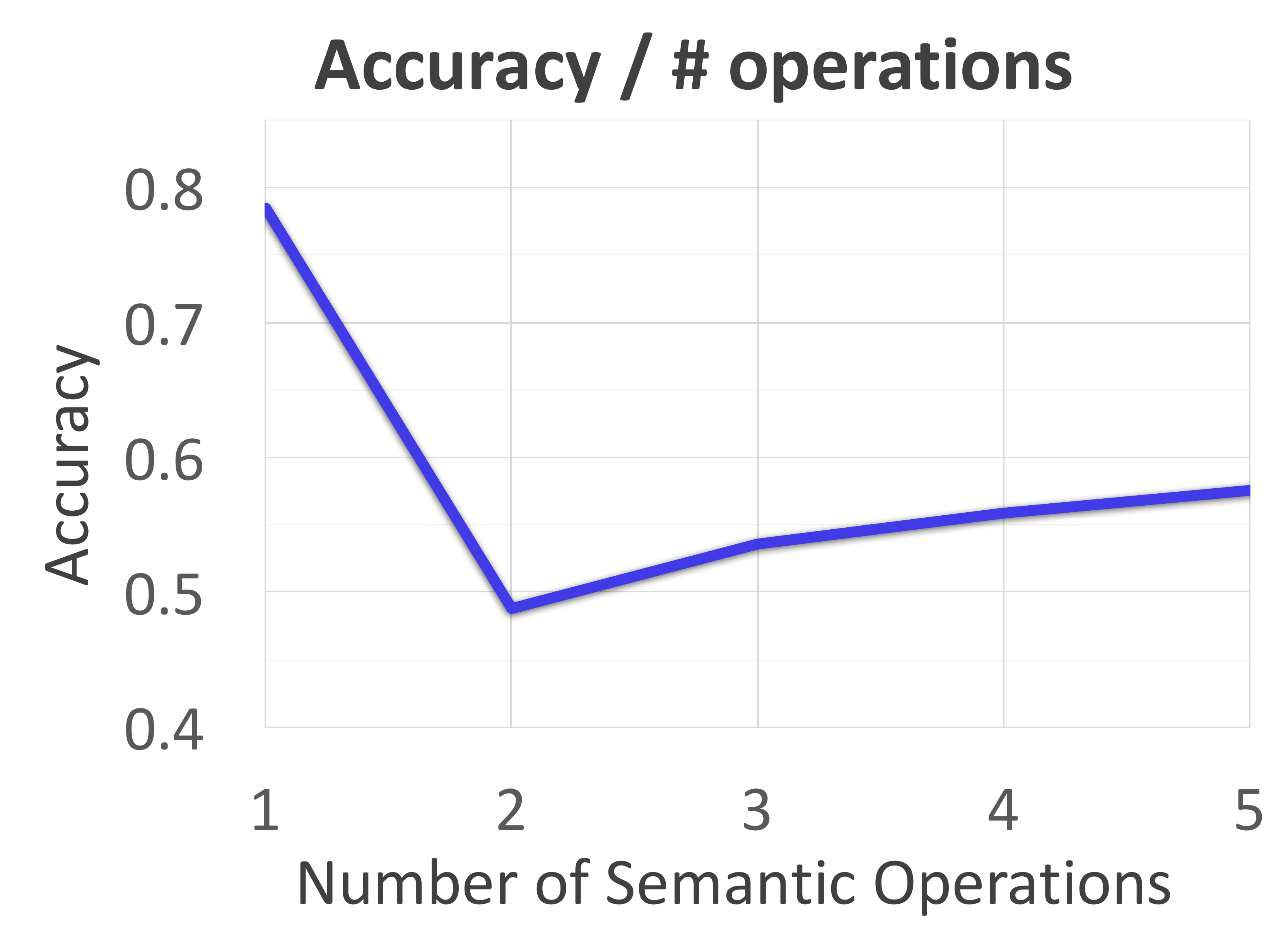}}
\hfill
\subfloat{\includegraphics[width=0.25\linewidth]{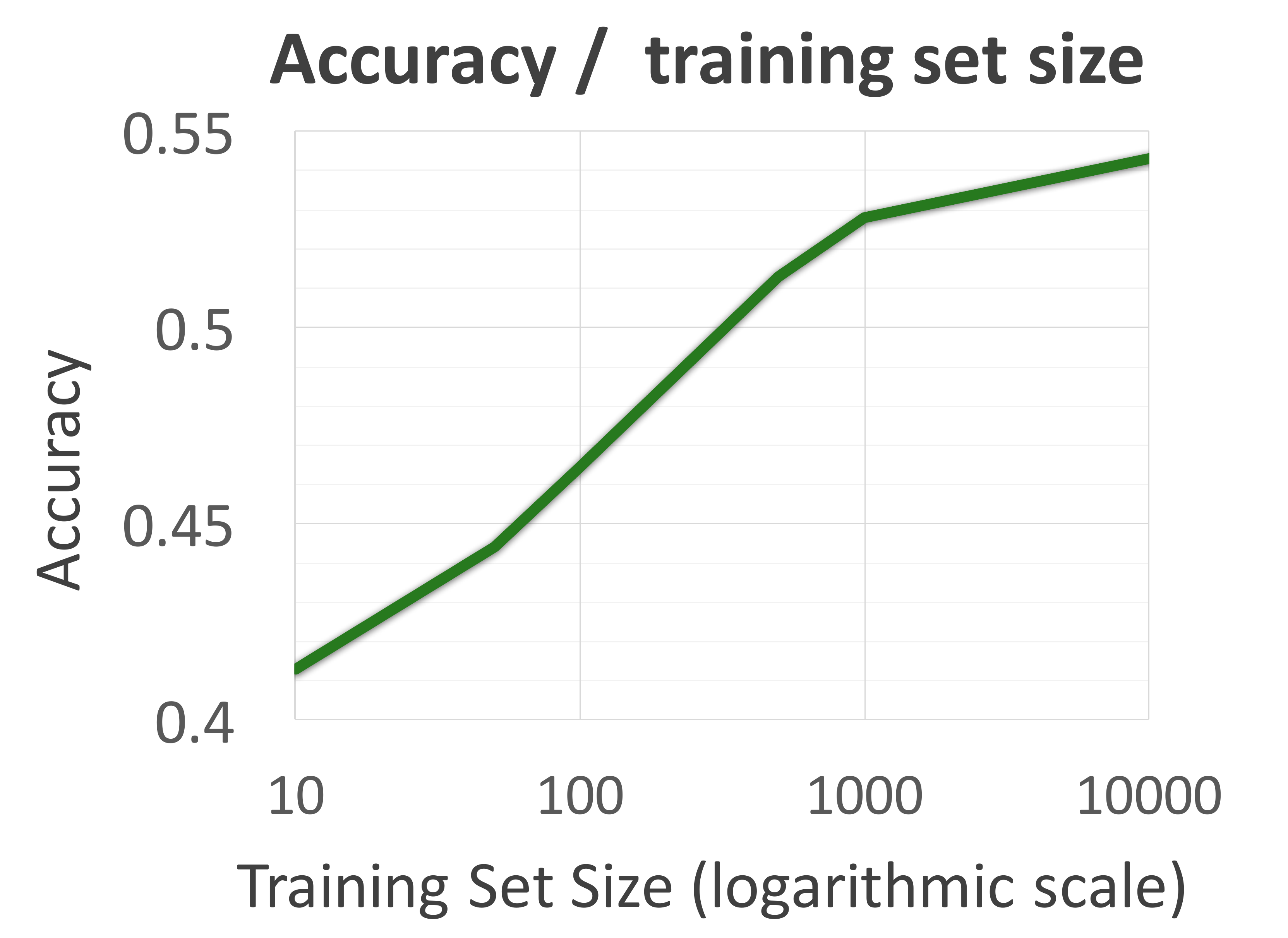}}
\hfill
\subfloat{\includegraphics[width=0.25\linewidth]{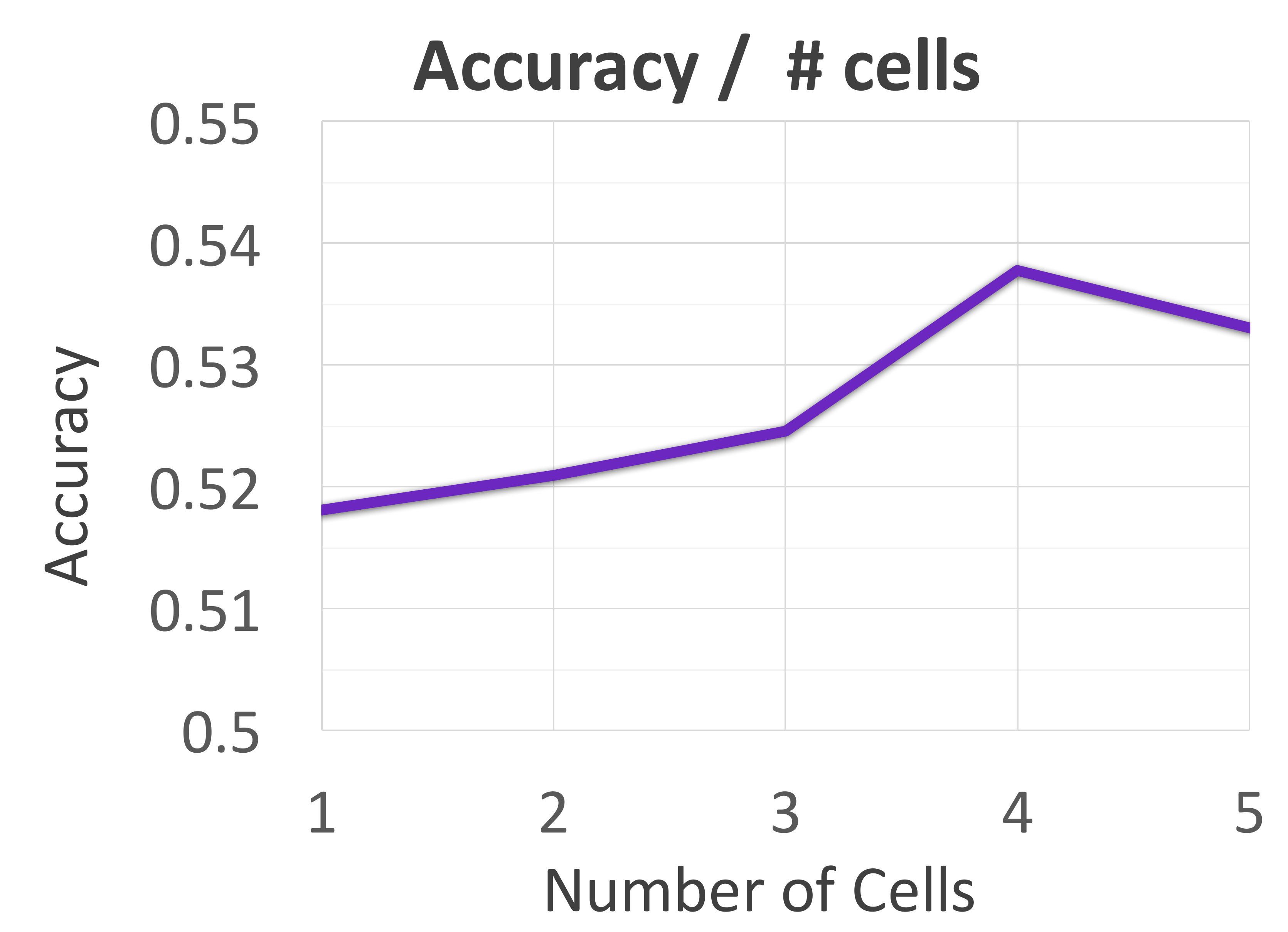}}
\caption{From left to right: \textbf{(1)} Accuracy as a function of textual question length -- the number of words in the question. \textbf{(2)} Accuracy as a function of semantic question length -- the number of operations in its functional program. \textbf{(3)} Performance as a function of the subset size used for training, ranging from 10K to 10M. \textbf{(4)} Accuracy for different lengths of MAC networks, suggesting that indeed GQA questions are compositional. } 
\label{fig:plots}
\end{figure}
\end{minipage}\qquad
\end{center}

We have used a sigmoid-based classifier and trained all models using Adam \cite{adam} for 15 epochs, each takes about an hour to complete. For MAC \cite{mac}, we use the official authored code available online, with 4 cells. For BottomUp \cite{bottomup}, since the official implementation is unfortunately not publicly available, we re-implemented the model, carefully following details presented in \cite{bottomup, tips}. To ensure the correctness of our implementation, we have tested the model on the standard VQA dataset, achieving 67\%, which matches the original scores reported by Anderson \etal \cite{bottomup}.



\section{Further Diagnosis}
\label{sec:exps2}

Following section 4.2 (main paper), and in order to get more insight into models' behaviors and tendencies, we perform further analysis of the top-scoring model for the GQA dataset, MAC \cite{mac}. The MAC network is a recurrent attention network that reasons in multiple concurrent steps over both the question and the image, and is thus geared towards compositional reasoning as well as rich scenes with several regions of relevance. 

We assess the model along multiple axes of variation, including question length, both textually, \ie number of words, and semantically, \ie number of reasoning operations required to answer it, where an operation can be \eg following a relation from one object to another, attribute identification, or a logical operation such as \textit{or}, \textit{and} or \textit{not}. We provide additional results for different network lengths (namely, cells number) and varying training-set sizes, all can be found in \figref{plots}.

Interestingly, question textual length correlates positively with the model accuracy. It may be the case that longer questions reveal more cues or information that the model can exploit, potentially sidestepping direct reasoning about the image. However, question semantic length has the opposite impact as expected: 1-step questions are particularly easy for models than the compositional ones which involve more steps. 



\begin{figure}
\vspace*{65mm}
\centering
\subfloat{\includegraphics[width=0.75\linewidth]{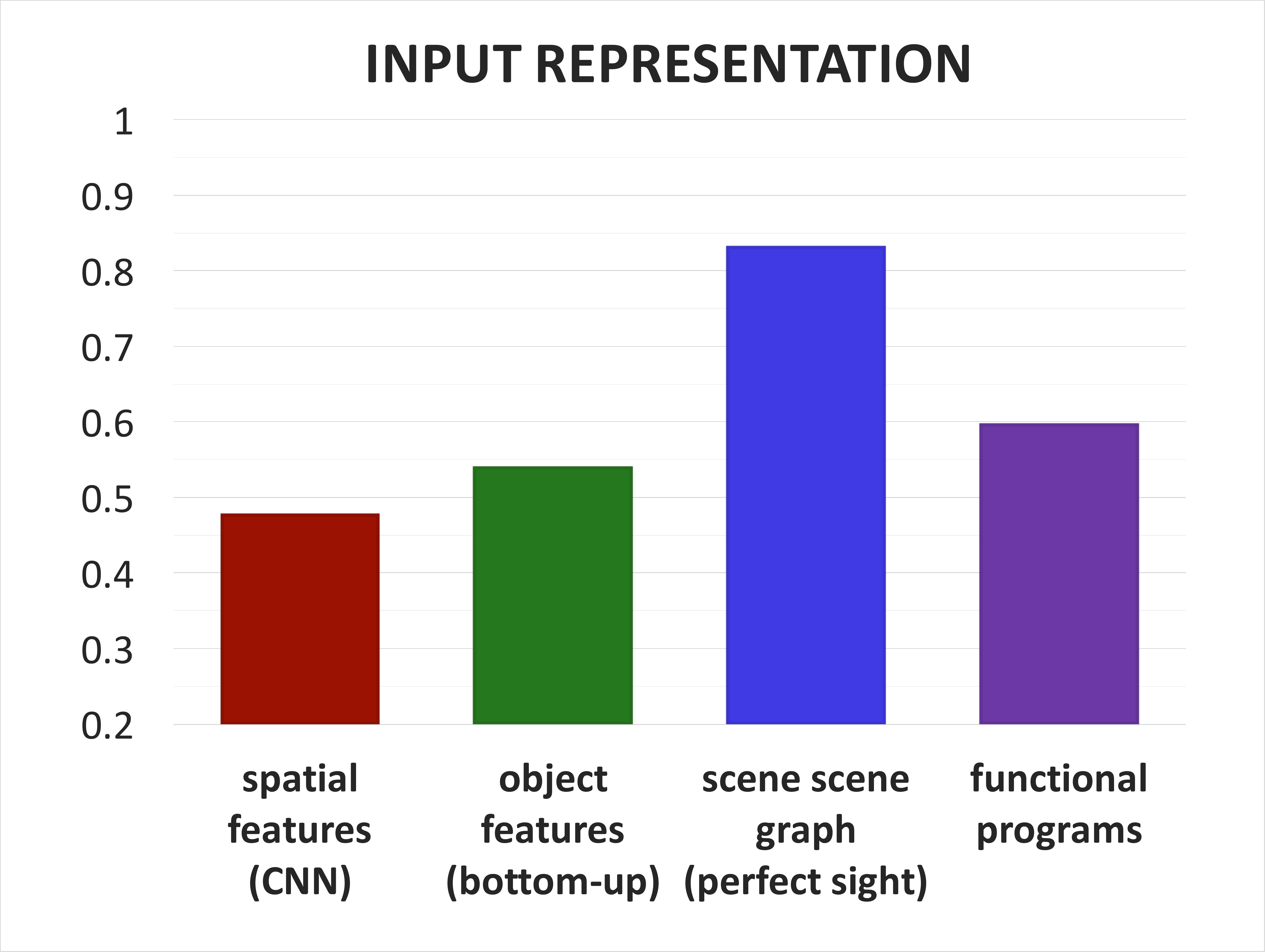}}
\caption{Performance as a function of the input representation. We encode the scenes through three different methods: spatial features produced by a standard pretrained CNN, object-based features generated by a faster R-CNN detector, and direct embedding of the scene graph semantic representation, equivalent to having perfect sight. We further experiment with both textual questions as well as their counterpart functional programs as input. We can see that the more semantically-imbued the representations get, the higher the accuracy obtained.}
\subfloat{\includegraphics[width=0.55\linewidth]{imgs/semcakeNew.pdf}}
\caption{Distribution of GQA questions semantic length (number of computation steps to arrive at the answer). We can see that most questions require about 2-3 reasoning steps, where each step may involve tracking a relation between objects, an attribute identification or a logical operation.}
\label{plotss}
\end{figure}


\clearpage

\begin{center}
\vspace*{-13mm}
\centering
\begin{minipage}[b]{1.0\textwidth}
\centering
\begin{figure}[H]
\hfill
\includegraphics[width=0.9\linewidth]{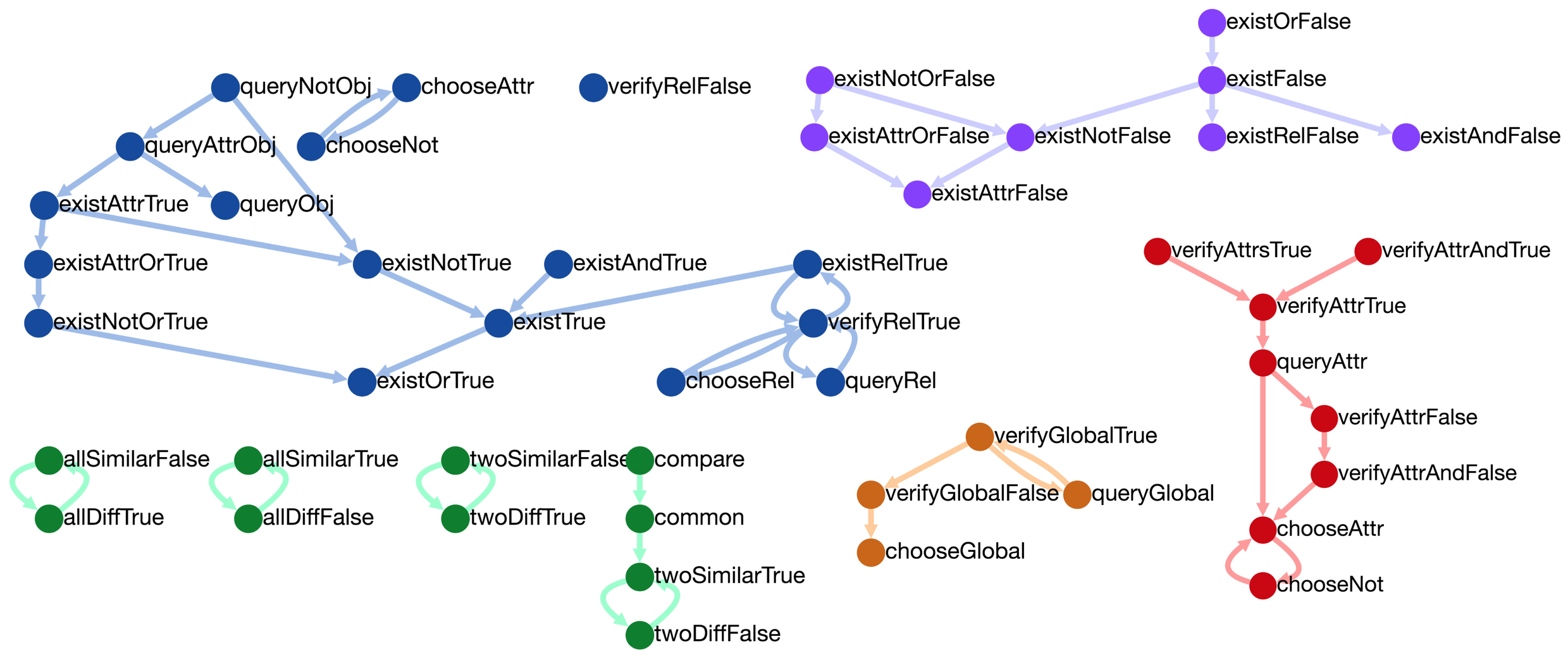}
\hfill
\vspace*{-1.7mm}
\caption{Entailment relations between different question types. In section 3.3 (main paper) we discuss the entailment and equivalences between questions. Since every question in the dataset has a matching logical representation of the sequence of reasoning steps, we can formally compute all the entailment and equivalence relations between different questions. Indeed, a cogent and reasonable learner should be consistent between its own answers, \eg should not answer ``red" to a question about the color of an object it has just identified as blue. Some more subtle relations also occur, such as those involving relations, \eg if \texttt{X} is above \texttt{Y}, than \texttt{Y} is below \texttt{X}, and \texttt{X} is not below \texttt{Y}, etc. \figref{entail} shows all the logical relations between the various question types. Refer to table \ref{table3} for a complete catalog of the different types. Experiments show that while people excel at consistency, achieving the impressive 98.4\%, deep learning models perform much worse in this task, with 69\% - 82\%. These results cast a doubt about the reliability of existing models and their true visual understanding skills. We therefore believe that improving their skills towards enhanced consistency and cogency is an important direction, which we hope our dataset will encourage.}

\label{fig:entail}
\end{figure}
\end{minipage}\qquad
\end{center}


\noindent\begin{minipage}[b]{1\textwidth}
\begin{multicols}{2}

We can further see that longer MAC networks with more cells are more competent in performing the GQA task, substantiating its increased compositionality. Other experiments show that increasing the training set size has significant impact on the model's performance, as found out also by Kafle \etal \cite{surveyMotivationStarve}.  Apparently, the training set size has not reached saturation yet and so models may benefit from even larger datasets. 

Finally, we have measured the impact of different input representations on the performance. We encode the visual scene with three different methods, ranging from standard pretrained CNN-based spatial features, to object-informed features obtained through faster R-CNNs detectors \cite{frcnn}, up to even a ``perfect sight" model that has access to the precise semantic scene graph through direct node and edge embeddings. As \figref{plots} shows, the more high-level and semantic the representation is, the better are the results. 

On the question side, we explore both training on the standard textual questions as well as the semantic functional programs. MAC achieves 53.8\% accuracy and 81.59\% consistency on the textual questions and 59.7\% and 85.85\% on the programs, demonstrating the usefulness and further challenge embodied in the former. It is also more consistent  Indeed, the programs consist of only a small operations vocabulary, whereas the questions use both synonyms and hundreds of possible structures, incorporating probabilistic rules to make them more natural and diverse. In particular, GQA questions have sundry subtle and challenging linguistic phenomena such as long-range dependencies, absent from the canonical programs. The textual questions thus provide us with the opportunity to engage with real, interesting and significant aspects of natural language, and consequently foster the development of models with enhanced language comprehension skills.

\section{Comparison between GQA and VQA 2.0}
We proceed by performing a comparison with the VQA 2.0 dataset \cite{vqa2}, the findings of which are summarized in table \ref{table1}. Apart from the higher average question length, we can see that GQA consequently contains more verbs and prepositions than VQA (as well as more nouns and adjectives)

\end{multicols}
\end{minipage}

\clearpage

providing further evidence for its increased compositionality. Semantically, we can see that the GQA questions are significantly more compositional than VQA's, and involve variety of reasoning skills in much higher frequency (spatial, logical, relational and comparative).

\begin{table}
\scriptsize 
\begin{center}
\begin{tabular}{lcc}
\rowcolor{Blue1}
\textbf{Aspect} & \textbf{VQA} & \textbf{GQA} \\
Question length & 6.2 + 1.9 & 7.9 + 3.1 \\
\rowcolor{Blue2}
Verbs & 1.4 + 0.6 & 1.6 + 0.7 \\
Nouns & 1.9 + 0.9 & 2.5 + 1.0 \\
\rowcolor{Blue2}
Adjectives & 0.6 + 0.6 & 0.7 + 0.7 \\
Prepositions & 0.5 + 0.6 & 1 + 1 \\
\rowcolor{Blue2}
Relation questions & 19.5\% & 51.6\% \\
Spatial questions & 8\% & 22.4\% \\
\rowcolor{Blue2}
Logical questions & 6\% & 19\% \\
Comparative questions & 1\% & 3\% \\
\rowcolor{Blue2}
Compositional questions & 3\% & 52\% \\
\end{tabular}
\end{center}
\caption{A head-to-head comparison between GQA and VQA 2.0. The GQA questions are longer on average, and consequently have more verbs, nouns, adjectives and prepositions than VQA, alluding to their increased compositionality. In addition, GQA demands increased reasoning (spatial, logical, relational and comparative) and includes significantly more compositional questions.}
\label{table1}
\end{table}


Some VQA question types are not covered by GQA, such as intention (\textit{why}) questions or ones involving OCR or external knowledge. The GQA dataset focuses on factual questions and multi-hop reasoning in particular, rather than covering all types. Comparing to VQA, GQA questions are objective, unambiguous, more compositional and can be answered from the images only, potentially making this benchmark more controlled and convenient for making research progress on.

\section{Scene Graph Normalization}
\label{sec:sgn}
Our starting point in creating the GQA dataset is the Visual Genome Scene Graph annotations \cite{vg} that cover 113k images from COCO \cite{coco} and Flickr \cite{flickr}.\footnote{We extend the original Visual Genome dataset with 5k new hidden scene graphs collected through crowdsourcing.} The scene graph serves as a formalized representation of the image: each node denotes an \textbf{object}, a visual entity within the image, like a person, an apple, grass or clouds. It is linked to a bounding box specifying its position and size, and is marked up with about 1-3 \textbf{attributes}, properties of the object: e.g., its color, shape, material or activity. The objects are connected by \textbf{relation} edges, representing actions (verbs), spatial relations (prepositions), and comparatives. 
The scene graphs are annotated with free-form natural language. Our first goal is thus to convert the annotations into a clear and unambiguous semantic ontology. We begin by cleaning up the graph's vocabulary, removing stop words, fixing typos, consolidating synonyms and filtering rare or amorphous concepts. We then classify the vocabulary into predefined categories (e.g., \textit{animals} and \textit{fruits} for objects; \textit{colors} and \textit{materials} for attributes), using word embedding distances to get preliminary annotations, which are then followed by manual curation. This results in a class hierarchy over the scene graph's vocabulary, which we further augment with various semantic and linguistic features like part of speech, voice, plurality and synonyms -- information that will be used to create grammatically correct questions in further steps. Our final ontology contains 1740 objects, 620 attributes and 330 relations, grouped into a hierarchy that consists of 60 different categories and subcategories. Visualization of the ontology can be found in \figref{ontology}.

At the next step, we prune graph edges that sound unnatural or are otherwise inadequate to be incoporated within the questions to be generated, such as \textit{(woman, in, shirt)}, \textit{(tail, attached to, giraffe)}, or \textit{(hand, hugging, bear)}. We filter these triplets using a combination of category-based rules, n-gram frquencies \cite{ngram}, dataset co-occurrence statistics, and manual curation.

In order to generate correct and unambiguous questions, some cases require us to validate the uniqueness or absence of an object. Visual Genome, while meant to be as exhaustive as possible, cannot guarantee full coverage (as it may be practically infeasible). Hence, in those cases we use object detectors \cite{frcnn}, trained on visual genome with a low detection threshold, to conservatively confirm the object absence or uniqueness.

Next, we augment the graph with absolute and relative positional information: objects appearing within the image margins, are annotated accordingly. Object pairs for which we can safely determine positional relations  (e.g., one is to the left of the other), are annotated as well. We also annotate object pairs if they share the same color, material or shape. Finally, we enrich the graph with global information about the image location or weather, if these can be directly inferred from the objects it contains.

By the end of this stage, the resulting scene graphs have clean, unified, rich and unambiguous semantics for both the nodes and the edges.

\begin{figure}
\centering
\subfloat{\includegraphics[width=\linewidth]{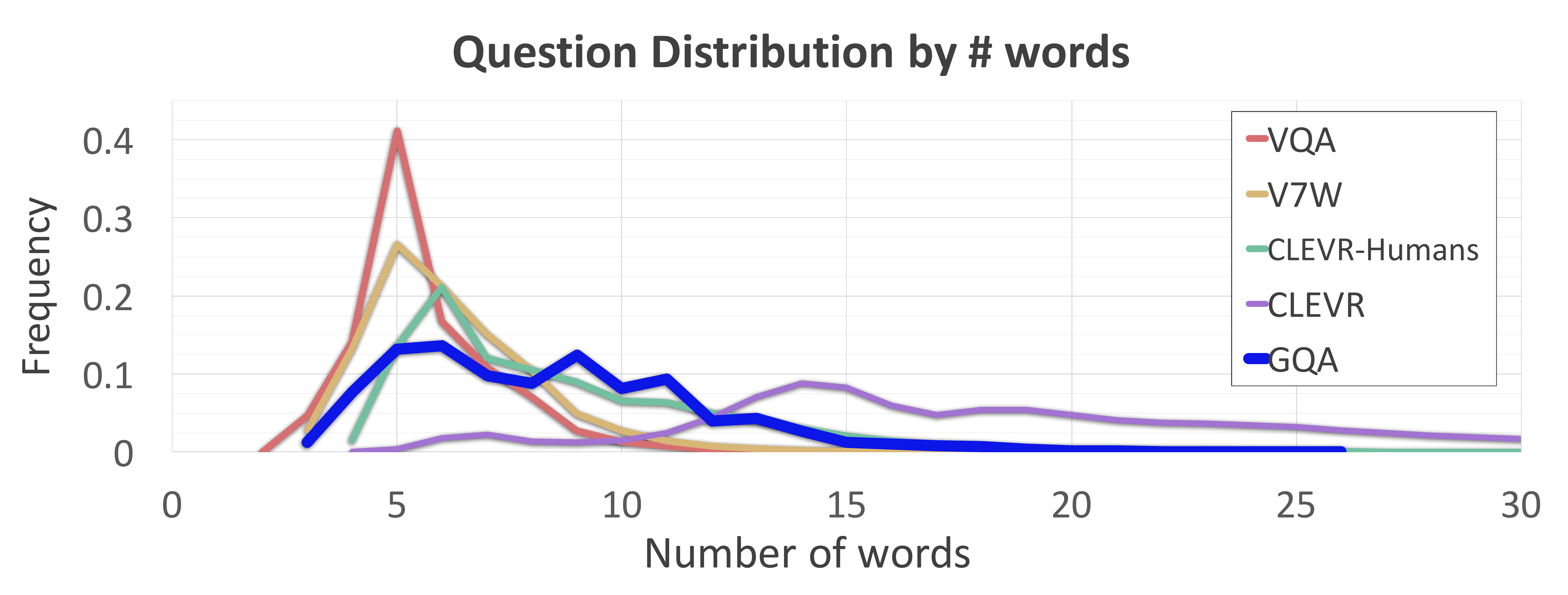}}
\caption{Question length distribution for Visual Question Answering datasets: we can see that GQA questions have a wide range of lengths and are longer on average than all other datasets except the synthetic CLEVR. Note that the long CLEVR questions tend to sound unnatural at times. }
\label{fig:distt}
\end{figure}

\clearpage

\begin{center}
\vspace*{-10mm}
\centering
\begin{minipage}[b]{1.0\textwidth}
\centering
\begin{figure}[H]
\centering
\includegraphics[width=0.826\linewidth]{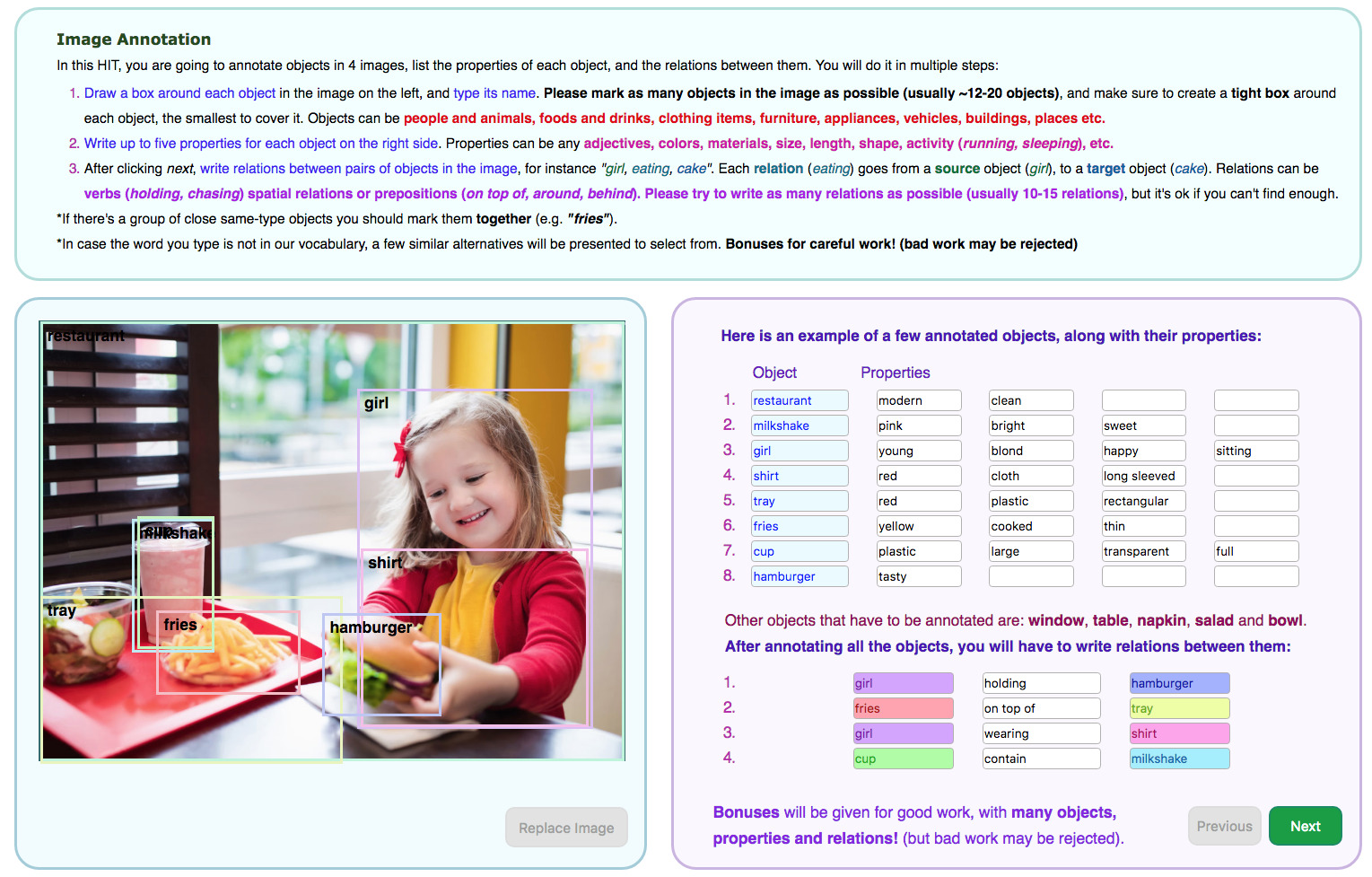}

\vspace*{5.1mm}

\includegraphics[width=0.826\linewidth]{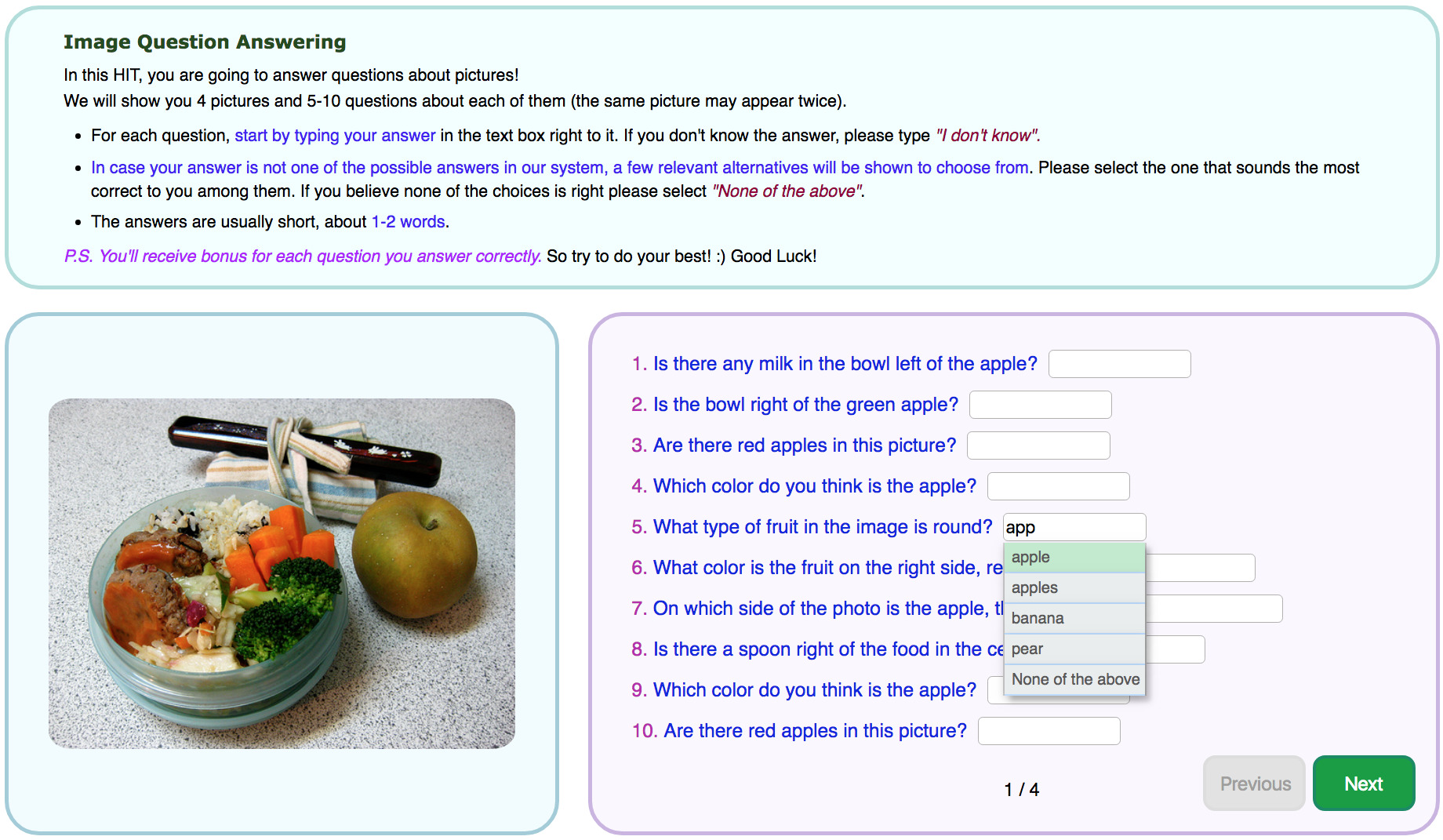}
\caption{The interfaces used for human experiments on Amazon Mechanical Turk. \textbf{Top}: Each HIT displays several images and asks turkers to list objects and annotate their corresponding bounding boxes. In addition, the turkers are requested to specify attributes and relations between the objects. An option to switch between images is also given to allow the turkers to choose rich enough images to work on. \textbf{Bottom}: Each HIT displays multiple questions and requires the turkers to respond. Since there is a closed set of possible answers (from a vocabulary with Approximately 1878 tokens), and in order to allow a fair comparison between human and models' performance, we give turkers the option to respond in unconstrained free-form language, but also suggest them multiple answers from our vocabulary that are the most similar to theirs (using word embedding distances). However, turkers are not limited to choose from the suggestions in case they believe none of the proposed answers is correct.} 
\end{figure}
\end{minipage}\qquad
\end{center}


\end{document}